\documentclass{article}
\usepackage[table]{xcolor} 
\usepackage{microtype}

\usepackage{hyperref}

\usepackage[accepted]{icml2025}

\usepackage{amsmath}
\usepackage{amssymb}
\usepackage{mathtools}
\usepackage{amsthm}

\usepackage[capitalize,noabbrev]{cleveref}

\usepackage{booktabs}       
\usepackage{amsfonts}       
\usepackage{overpic}

\usepackage{natbib}
\usepackage{amsmath}
\usepackage{amssymb}
\usepackage{pdfpages}
\usepackage{graphicx}
\graphicspath{ {./res/} }
\usepackage{caption}
\usepackage{subcaption}
\usepackage{multirow}
\usepackage{dirtytalk}
\usepackage{float} 

\usepackage{siunitx}

\newcommand{\AR}[1]{{\textcolor{red}{[\textbf{AR:} #1]}}}
\definecolor{cellblue}{RGB}{173,216,230}

\theoremstyle{plain}

\theoremstyle{definition}

\theoremstyle{remark}

\usepackage[textsize=tiny]{todonotes}

\icmltitlerunning{R3L: Relative Representations for Reinforcement Learning
}

\begin{document}

\twocolumn[
\icmltitle{R3L: Relative Representations for Reinforcement Learning}




\begin{icmlauthorlist}
\icmlauthor{Antonio Pio Ricciardi}{sap}
\icmlauthor{Valentino Maiorca}{sap,ista}
\icmlauthor{Luca Moschella}{sap}
\icmlauthor{Riccardo Marin}{tub}
\icmlauthor{Emanuele Rodolà}{sap}
\end{icmlauthorlist}

\icmlaffiliation{sap}{Sapienza, University of Rome}
\icmlaffiliation{tub}{University of Tübingen, Germany}
\icmlaffiliation{ista}{Institute of Science and Technology Austria (ISTA)}


\icmlcorrespondingauthor{Antonio Pio Ricciardi}{ricciardi@di.uniroma1.it}

\icmlkeywords{Machine Learning, ICML}

\vskip 0.3in
]



\printAffiliationsAndNotice{}  

\begin{abstract}


Visual Reinforcement Learning is a popular and powerful framework that takes full advantage of the Deep Learning breakthrough. It is known that variations in input domains (e.g., different panorama colors due to seasonal changes) or task domains (e.g., altering the target speed of a car) can disrupt agent performance, necessitating new training for each variation. Recent advancements in the field of representation learning have demonstrated the possibility of combining components from different neural networks to create new models in a zero-shot fashion. In this paper, we build upon relative representations, a framework that maps encoder embeddings to a universal space. We adapt this framework to the Visual Reinforcement Learning setting, allowing to combine agents components to create new agents capable of effectively handling novel visual-task pairs not encountered during training.
Our findings highlight the potential for model reuse, significantly reducing the need for retraining and, consequently, the time and computational resources required.

\end{abstract}
\section{Introduction}
\begin{figure*}[t]
    \centering
    \begin{subfigure}[b]{0.4\linewidth}
        \centering
        \includegraphics[width=1\linewidth]{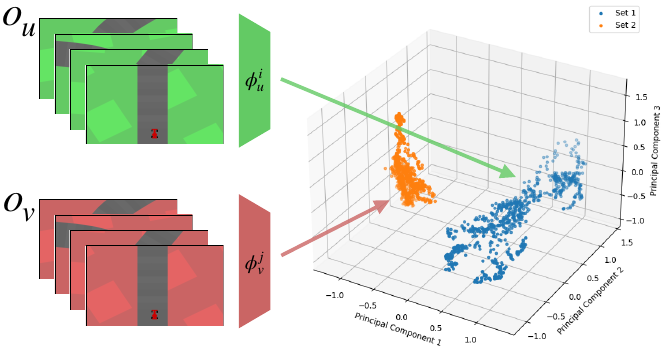}
        \label{subfig:abs-repr}
    \end{subfigure}%
    \hspace{2mm}
    \begin{subfigure}[b]{0.5\linewidth}
        \centering
        \includegraphics[width=1\linewidth]{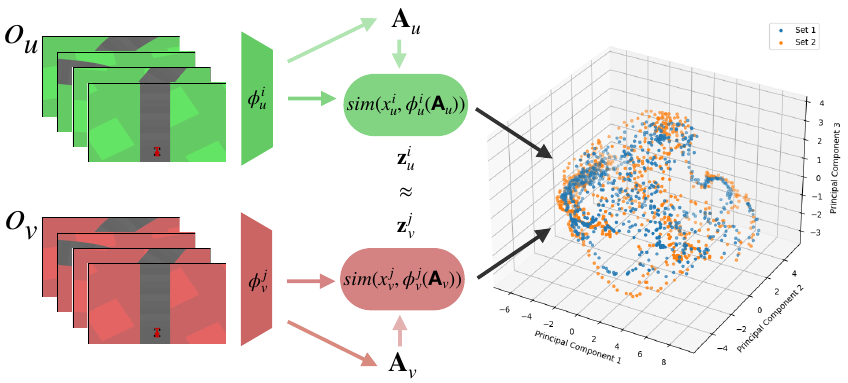}
        \label{subfig:rel-repr}
    \end{subfigure}
    \caption{\textbf{Left:} Standard (absolute) encoder outputs for the two visual variations (red/green). Notice how the embeddings occupy two distinct regions of the space.
    \textbf{Right:} R3L encoder outputs. Here, points from the two sets align nearly one‐to‐one, indicating closer correspondence in their learned representations.}
    \label{fig:abs-relrepr-architecture}
\end{figure*}

Reinforcement Learning (RL) drives some of the most prominent achievements in modern artificial intelligence. Its combination with deep learning enables superhuman performances in complex and articulated tasks like strategic games \cite{silver2016mastering, silver2017mastering}, showing micro and macro adaptability to settings with a wide variety of inputs  \cite{vinyals2019grandmaster}. 

Despite its success, Deep RL remains inaccessible to many due to its high computational cost. Training agents without labeled data requires millions of interactions, often taking weeks even for simple tasks. Additionally, these architectures are highly sensitive to training conditions; small changes, such as different random seeds, can lead to drastically different outcomes. Finally, agents tend to overfit to the environments they were trained on, making them ineffective when faced with shifts in visuals, tasks, or domains, often requiring training new agents from scratch.

Many approaches address domain shift {\em a priori} using domain randomization \cite{tobin2017domain} or data augmentation techniques \cite{hansen2021softda, yarats2021image, yarats2021drqv2, hansen2021stabilizing}. These methods promote invariance by training models to ``memorize'' irrelevant variations, but they often require longer training times and more complex architectures. Additionally, they rely on prior knowledge of the perceptual shifts that may occur during deployment.

Latent alignment techniques \cite{hansen2020self, yoneda2021invariance} instead focus on achieving feature invariance at deployment, assuming the task remains unchanged. This is done by collecting latent features during training as a reference for the agent's understanding of the task. When encountering a new visual domain, new latents are trained to match the distribution of those seen during training. Other methods impose training constraints to make neural components reusable \cite{devin2017learning, jian2023policy}. Recent advances in representation learning further demonstrate that neural components can be reused in a {\em zero-shot} fashion by projecting different models' latent spaces into a shared space \cite{Moschella2022-yf}, or by mapping the produced latent space from a model to another \cite{maiorca2023latent}.

In this work, we demonstrate how visual deep RL methods can be extended to enable zero-shot recombination of learned representations and skills across different environments. Standard RL methods struggle with generalization: for instance, an agent trained to drive on a racing track during spring would fail in summer due to changes in grass color, as its encoder maps observations to entirely different latent representations. This discrepancy necessitates retraining, limiting the reuse of learned policies.

To address this, we propose Relative Representations for Reinforcement Learning (R3L), which aligns latent spaces across visual and task domains, enabling neural component reuse without retraining. Unlike absolute feature mappings, which shift under distributional changes, R3L preserves relative relationships, ensuring compatibility between encoders and controllers trained in different settings.

\Cref{fig:abs-relrepr-architecture} illustrates this: (Left), standard RL methods produce non-comparable latent spaces, making transfer impossible. (Right), R3L unifies these spaces, allowing seamless recombination of learned components.

\paragraph{Contributions}
We present R3L, a method that, leveraging advances in representation learning ~\citep{Moschella2022-yf} ,enables encoders and controllers trained in different settings to be combined without additional fine-tuning, adapting to both perceptual shifts and task variations, by unifying the latent spaces produced by the encoders. Our experiments across diverse environments highlight underlying shared structures in RL, revealing new theoretical and practical directions for modular learning and compositional generalization.

In summary, our work explores emerging patterns in representations learned across different environments. We propose viewing end-to-end trained models as a combination of two components, leveraging this insight to introduce a novel zero-shot stitching procedure. This enables policy components trained in one environment variation to transfer seamlessly to another. As demonstrated in \Cref{sec:experiments}, our method generalizes across observation shifts (e.g., background changes, camera angles) and variations in task objectives (e.g., different goals or action spaces).



\section{Related Work}

\paragraph{Model Stitching}
The concept of {\em model stitching} has been explored to analyze the similarity of latent spaces across models. For instance, stitching layers as discussed in \cite{Lenc2014-gy, Bansal2021-oj, Csiszarik2021-yi} provide a metric for measuring similarity.
Recently, model stitching has been extended to facilitate model reuse by integrating parts from multiple networks. Instead of designing compatible and reusable components, which can complicate the model architecture \citep{Gygli2021-qb, https://doi.org/10.48550/arxiv.2208.03861}, these approaches enable {\em zero-shot stitching} between neural networks \cite{Moschella2022-yf, norelli2022b, maiorca2023latent, cannistraci2023bricks}. They assume a semantic correspondence between training distributions to unify representations within a shared space or estimate a direct translation between them.


\paragraph{RL Modular Agents} Transferring agent policies across visual or dynamic environment variations remains an open challenge \cite{zhu2023transfer, mohanty2021measuring, jian2021adversarial}. Various approaches have explored reusing components such as value functions \cite{tirinzoni2018transfer, liu2021learning}, policies \cite{fernandez2006probabilistic, devin2017learning}, parameters \cite{killian2017robust, finn2017model}, and features \cite{barreto2017successor}. Similarly, \citet{zhang2020learning} minimize bisimulation metrics to train encoders invariant to visual distractors or small input differences by combining reward signals with latent space distance.

Another strategy is designing agents as compositions of multiple modules. Modular RL follows a ``sense-plan-act'' paradigm \cite{karkus2020beyond}, addressing sample inefficiency by integrating specialized modules that solve different sub-problems \cite{mendez2022modular}. When combined, these modules enable agents to tackle more complex tasks \cite{russell2003q, simpkins2019composable}. In many cases \cite{wolf2023augmented}, prior knowledge and external information guide the process. Here, a module functions as a policy for a sub-task, while a neural network acts as a planner, selecting the appropriate policy in real time \cite{sutton2011horde, goyal2019recurrent}. However, a key limitation of modular approaches is their reliance on specific architectures and careful manual design.




\paragraph{Latent alignment for RL} Another line of RL research focuses on invariances in latent space, based on the assumption that for certain input variations, \say{the task remains the same, and so the agent experiences internally should also remain the same} \cite{yoneda2021invariance}. For example, PAD \cite{hansen2020self} trains an inverse dynamics model in a supervised manner, while ILA \cite{yoneda2021invariance} fine-tunes to align prior distributions under visual changes without requiring paired image data. However, both methods address only visual variations, ignoring task changes.

In contrast, \cite{devin2017learning} explores task-robot stitching using interconnected modules but requires limiting the number of hidden units for tractability. Their approach combines trajectory optimization and supervised learning within a global neural network policy, using regularization to enforce task invariance and prevent overfitting to training robot-task pairs. More recently, \cite{jian2023policy} demonstrated policy stitching for a robotic arm using the relative representation framework \cite{Moschella2022-yf}, which, however, is constrained to low-dimensional inputs.

\paragraph{Our Positioning} Building on \cite{Moschella2022-yf}, we frame our work within representation learning for RL, focusing on zero-shot stitching of neural models trained under different conditions. We achieve this by projecting policy encoders into a shared latent space, enabling seamless communication between them. Using relative representations, controllers can interpret encoders' latent spaces without fine-tuning, provided they share the same space. Unlike \citet{jian2023policy}, we operate on high-dimensional image features rather than low-dimensional states. This approach enables stitching models trained on diverse visual and task variations, creating new policies for unseen visual-task combinations in a zero-shot manner.

\newcommand{\enc}[0]{\phi}
\newcommand{\con}[0]{\psi}
\newcommand{\encmap}[2]{\mathcal{O}_{#1}^{#2} \mapsto \mathcal{X}_{#1}^{#2}}
\newcommand{\conmap}[3]{\mathcal{X}_{#1}^{#2} \mapsto \mathcal{A}_{#3}}
\newcommand{\ours}[0]{Trasl. \textbf{(Ours)}}

\section{Method}
\paragraph{Context} We formally model an RL problem as a Markov decision process $\mathcal{M} = ( \mathcal{S, A}, \mathcal{O}, R, P, \gamma )$, where
$\mathcal{S}$ defines the set of states, $\mathcal{A}$ the set of actions, $o \in \mathcal{O}$ the observation produced as input for the agents, $P : \mathcal{S \times A} \mapsto \mathcal{S}$ the probability distribution $P(\mathbf{s'} | \mathbf{s, a})$ of transitioning to state $\mathbf{s'}$ upon executing action $\mathbf{a}$ in state $\mathbf{s}$, $\mathcal{R} : \mathcal{S \times A} \mapsto \mathcal{R}$ the reward function, and $\gamma$ the discount factor that reduces the importance of rewards obtained far in the future. The agent’s behavior is dictated by a policy $\pi : \mathcal{O} \rightarrow \mathcal{A}$ that receives an observation and selects an action at each state, and is trained to maximize the discounted long-term returns $\mathbb{E} [\sum_{i=0}^{\infty} \gamma^i \mathcal{R}(s_i, a_i)]$.

\paragraph{Environments Variations} 
We study the effects of variations in training conditions and agent behavior. To formalize ``variation'', we redefine environments as follows. 

An environment $\mathcal{M}_u^i = (\mathcal{O}_u, T_i)$ produces \emph{observations} $o_u \in \mathcal{O}_u$ and requires solving a \emph{task} $T_i : \mathcal{S}_i\times \mathcal{A}_i\times \mathcal{R}_i \times{P}_i \mapsto \mathcal{R}_i$. Observation distributions $\mathcal{O}_u$ differ when they exhibit significant changes (e.g., background color, camera perspective). Task variations affect agent behavior through transition dynamics, internal states, action spaces, and reward functions. Since task $T_i$ information is not explicitly provided, agents must infer it from reward signals.

We train agents separately on different visual-task variations. During testing, we evaluate environments on unseen visual-task combinations, meaning no agent has been trained end-to-end on $\mathcal{M}_u^j$ to solve $T_j$ with observations $O_u$.

\paragraph{Modular Agents} 
The standard practice to obtain a policy $\pi_u^i$ is to end-to-end train a neural network on environment $\mathcal{M}_u^i$.
The network can be seen as a composition of two functions: (i) an {\em encoder} $\enc_u^i : \encmap{u}{i}$ trained on an environment with an observation space $\mathcal{O}_u$ to produce a latent representation $\mathbf{x}_u^i$; and (ii) a {\em controller} $\con_u^i: \conmap{u}{i}{i}$ trained to act on task $T_i$ given the latent representation coming from the encoder. Given an observation $o_{u}$, $\pi_u^i$ can be defined by composition:
\begin{equation}
    \pi_u^i(o_{u}) = \con_u^i[\enc_u^i(o_{u})] = \con_u^i(\mathbf{x}_{u}^i)\,.
\end{equation}

\paragraph{Latent representation} Consider a second environment $\mathcal{M}_v^j = ( \mathcal{O}_v, \mathcal{T}_j )$, where the observations $\mathcal{O}_v$ differ from $\mathcal{O}_u$ just for a visual variation (e.g., different grass color in CarRacing), and a trained policy $\pi_v^j$. Given two corresponding observations $o_{u} \in O_u$ and $o_{v} \in O_v$, the latent representations produced by the respective encoders are different:
\begin{equation}
\begin{split}
    \enc_u^i(o_{u}) \neq \enc_v^j(o_{v}) \quad \text{and therefore} \quad \mathbf{x}_{u}^i \neq \mathbf{x}_{v}^j\,.
\end{split}
\end{equation}
We next describe how to unify learned latent representations by leveraging their emerging similarities. This enables {\em zero-shot stitching} between encoders and controllers by projecting their latent spaces into a shared space (\cref{sec:method-relative}).

\subsection{Relative representations}\label{sec:method-relative}
The core idea of relative representations \citep{Moschella2022-yf} is to encode latent space elements not as their original (\textit{absolute}) embeddings but in relation to a set of selected samples, called \emph{anchors}. Namely, given an observation $o_{u}$ with latent representation $\enc_u^i(o_{u}) = \mathbf{x}_{u}^i$, the \emph{relative representation} $\mathbf{z}_{u}^i$ is computed based on its similarity to the anchor set $\mathbf{A}$:
\begin{align}
\centering
     \mathbf{z}_{u}^i &= sim(\enc_u^i(o_{u}), \enc_u^i(\mathbf{A}_u)) \\&= sim(\mathbf{x}_{u}^i, \enc_u^i(\mathbf{A}_u)) \\&= 
    [sim(\mathbf{x}_{u}^i, \enc_u^i(\mathbf{A}_u^{(0)})), sim(\mathbf{x}_{u}^i, \enc_u^i(\mathbf{A}_u^{(1)})),\\& \dots sim(\mathbf{x}_{u}^i, \enc_u^i(\mathbf{A}_u^{(d)}))],
\end{align}
where $d$ is the dimension of the latent space and one of the controller's inputs. Per the original paper, we select the cosine similarity as $sim$ function. However, the original anchor selection process assumes the availability of an offline dataset to sample $\mathbf{A}$ from; we generalize this assumption to the online RL setting in \Cref{sec:datacollection}.


\paragraph{Intuition} This relative representation ignores absolute latent positions, which are heavily shaped by the agent's training process, and instead captures relations between observations. While encoders may produce different absolute latent representations due to environment variations, they are expected to yield {roughly similar relative representations}:
\begin{align}
    sim(\enc_u^i(o_u), \enc_u^i(\mathbf{A}_u)) & \approx sim(\enc_v^j(o_{v}), \enc_v^j(\mathbf{A}_v))\\
    sim(\mathbf{x}_{u}^i, \enc_u^i(\mathbf{A}_u)) & \approx sim(\mathbf{x}_{v}^j, \enc_v^j(\mathbf{A}_v))\\
    \mathbf{z}_{u}^i & \approx \mathbf{z}_{v}^j
\end{align}
Thus, we \emph{train the controller directly on the relative spaces} to produce a universal controller, reusable across a variety of settings.

\paragraph{Module stitching}\label{sec:stitchingdef} We exploit the similarity between latent relative representations of different agents to perform zero-shot stitching. For example, if we are given policies $\pi_u^i$ and $\pi_v^j$ trained with relative representations to act on $\mathcal{M}_u^i = ( \mathcal{O}_u, \mathcal{T}_i ) $ and $\mathcal{M}_v^j = ( \mathcal{O}_v, \mathcal{T}_j )$ respectively, we can combine the encoder $\enc_u^i$ from $\pi_u^i$ and the controller $\con_v^j$ from $\pi_v^j$ to create a new policy $\hat{\pi}_u^j$ that can act on $\mathcal{M}_u^j = ( \mathcal{O}_u, \mathcal{T}_j )$:
\begin{equation}
    \hat{\pi}_u^j(o_{u}) = \con_v^j(\mathbf{z}_{u}^i)\,.
\end{equation}
This can be done because encoders $\enc_u^i$ and $\enc_v^j$ produce similar latent spaces, therefore controllers $\con_u^i$ and $\con_v^j$ are trained on similar representations.
Throughout this paper, we will use encoders and policies trained using PPO.

\paragraph{R3L}
RL training is highly unstable due to the constantly shifting objective, where learning targets depend on the agent’s own evolving policy and representation (
\citep{sutton2018reinforcement, mnih2015human, lillicrap2015continuous},
Using anchors adds even more instability, due to the fact that we are updating the encoder with which we compute the new latents for the anchors, and therefore we keep moving the landmarks on which the controller is trained.

Thus, we propose a simple yet effective solution to stabilize training. We use the exponential moving average ({EMA}) to avoid changing the anchors too quickly, preserving part of the old anchors:
\begin{equation}
    \phi (\textbf{A}_t) = \alpha \phi (\textbf{A}_t) + (1 - \alpha) \phi (\textbf{A}_{t-1})\,,
\end{equation}
where -- simplifying our notation and forgetting about background variations for a moment -- $\textbf{A}_{t-1}$ are the previous anchors and $\textbf{A}_t$ are the new ones, as output by the updated encoder. We update the anchors at every training step.

We selected $\alpha = 0.999$ for all experimental trials. Subsequent analyses in the experiments section compare training outcomes across different $\alpha$ values, substantiating the choice of 0.999 as the optimal parameter.

\subsection{Data Collection}\label{sec:datacollection}
 
Prior work on latent space alignment \citep{Moschella2022-yf, jian2023policy, maiorca2023latent} relies on supervision from (\emph{parallel anchors}), training data samples with partial correspondence across domains. However, online RL lacks such predefined training data. To overcome this, we assume a mapping function that translates observations from one environment $\mathcal{O}_{u}$ to a target $\mathcal{O}_{v}$. This mapping can be estimated through manual annotation, replaying action sequences, or directly modeling transformations in the observation space (e.g., pixel space).
%
%

\begin{figure}[!t]
        \centering        \includegraphics[width=0.4\textwidth]{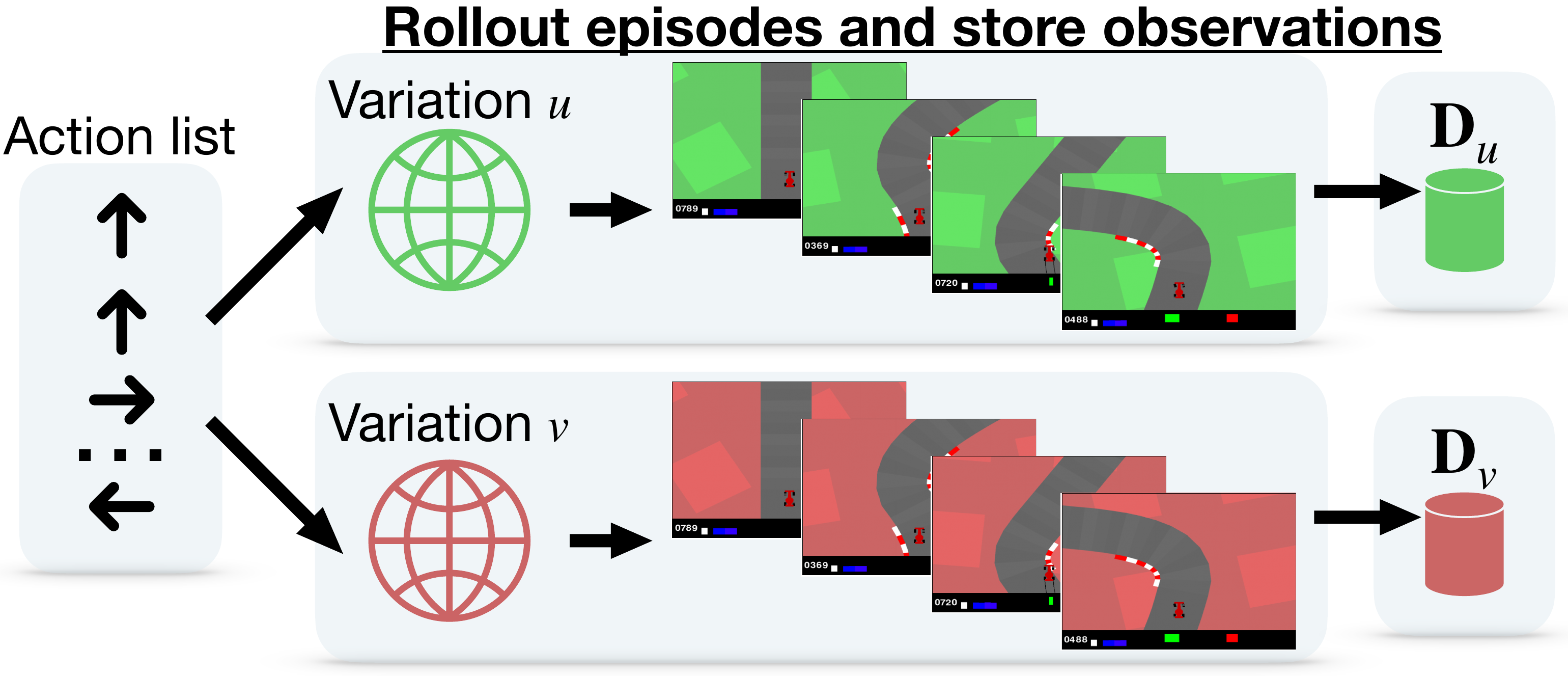}
        \caption{Data collection pipeline. We can either use the same sequence of actions on two different env variations, or apply a transformation on top of another set.}
        \label{fig:data-collection}
\end{figure}

In our experiments, we obtain parallel samples either by translating observations in pixel space when the visual variation is well-defined or by replaying identical action sequences in both environments, ensuring determinism with the same random seed (\Cref{fig:data-collection}). Exploring alternative methods for approximating observation translation across environments is left for future work.

\section{Experiments}\label{sec:experiments}
In this section, we present both qualitative and quantitative evaluations of R3L. We begin by examining the latent‐space similarities produced by the relative representations framework, in \Cref{sec:exp-analysis}. In \Cref{sec:exp-end-to-end-performance}, we compare the training curves of the models trained using R3L against those of the standard (absolute) models, and also evaluate their performance without any stitching. Next, in \Cref{sec:zero-shot-stitching}, we quantitatively compare the zero‐shot stitching performance between R3L and a naive baseline (i.e., no latent communication). Finally, in \Cref{sec:finding-ema}, we describe how we select the best $\alpha$ value for the exponential moving average.


\paragraph{Notation}
We refer to standard, end-to-end policies using \textit{absolute} representations as \emph{E. Abs}, and to \emph{E. R3L} for end-to-end policies trained using relative representations.
We will use \emph{S. Abs}, and to \emph{S. R3L} to instead refer to policies created with zero-shot-stitching through relative representations, where encoders and controllers variations can include seed, background colors and tasks.
Unlike end-to-end models, stitched agent modules come from encoders and controllers trained independently and assembled as detailed in \Cref{sec:method-relative}. We also include \emph{E. Multicolor} and \emph{S. Multicolor} to indicate models trained with background color changing after every reset.

\paragraph{Environments}
For the following experiments, we consider the CarRacing \citep{klimov2016carracing} environment, which simulates driving a car from a 2D top-down perspective around a randomly generated track, focusing on speed while staying on the track. It provides RGB image observations, and uses a discretized action space with five options: steer left, steer right, accelerate, brake, and idle. The agent earns rewards for progress and penalties for going off-track. We modified the environment to enable visual changes (e.g, grass color or camera zoom) and task alterations (e.g., speed limits or different action spaces).
The possible visual variations are: background (grass) colors \textit{green}, \textit{red}, \textit{blue} and \textit{far camera zoom}, while tasks are divided in: \textit{standard} and \textit{slow} car dynamics and different action spaces, such as \textit{scrambled}, which use a different action space and therefore a different output order for the car commands, and \textit{no idle}, which removes the \say{idle} action.
We also test R3L with some environments of the Atari game suite. Please refer to \Cref{appendix:atari} for the implementation and additional tests with this environment.

\paragraph{Training procedures}
We train policies using the PPO implementation provided in the CleanRL library \citep{huang2022cleanrl} with default hypermarameters for both absolute models and R3L.

\paragraph{Zero-Shot Stitching Procedure.}
In \Cref{sec:method-relative} we outlined the methodologies for stitching modules together using relative representations. 
We consider the \textit{encoder} to be the group of convolutional layers up to the first flatten layer, while the \textit{controller} is everything that comes immediately after, that is a succession of linear layers and activation functions.
We chose this division primarily because we stack frames at the embedding level for visual observations. Since anchors are collected as single frames, we can independently embed each frame to the relative space, and then stack those representations before passing them on to the controller, that requires stacked embeddings as input.

Once diverse policies are trained under various visual and task variations, at test time we can generate new policies by assembling independently trained encoders and controllers through zero-shot stitching, to play on new visual-task variation never-before-seen \textit{together} during training.
The zero-shot stitching performance evaluation is always on visual-task variation not seen during training.
%
%
It is crucial to select encoders and controllers that correspond to the specific visual or task variations they were trained on. For instance, when operating within an environment featuring a green background, an encoder trained on that specific visual variation should be utilized. Similarly, for tasks that involve driving a car at low speeds, a controller trained for that specific driving condition must be employed.

\paragraph{Computing relative representations} We use the cosine distance to compute the relative representations. Please note that relative representations can be computed starting from already existing models.
However, we need to train the entire pipeline so that controllers are able to interpret the new \emph{universal} representations. In \Cref{sec:exp-analysis} we compute relative representations starting from an absolute model to show that spaces can be compatible.
In \Cref{sec:exp-end-to-end-performance} we compare the performance of models trained end-to-end with absolute and relative representations.

\begin{figure}[t!]
    \centering
    \begin{subfigure}[b]{0.6\linewidth}
        \centering
        \includegraphics[width=\linewidth]{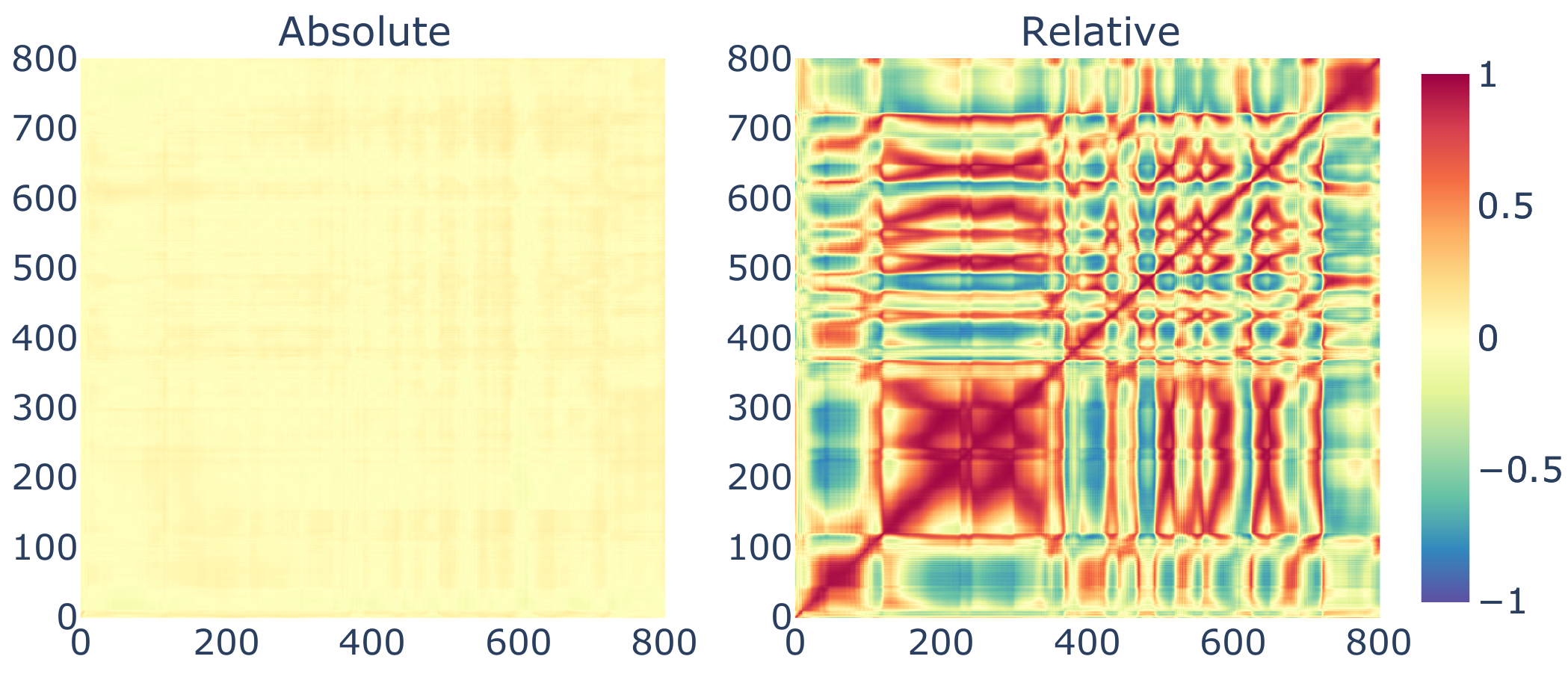}
        \caption{}
        \label{fig:matrix-sim}
    \end{subfigure}%
    \begin{subfigure}[b]{0.4\linewidth}
        \centering
        \includegraphics[width=\linewidth]{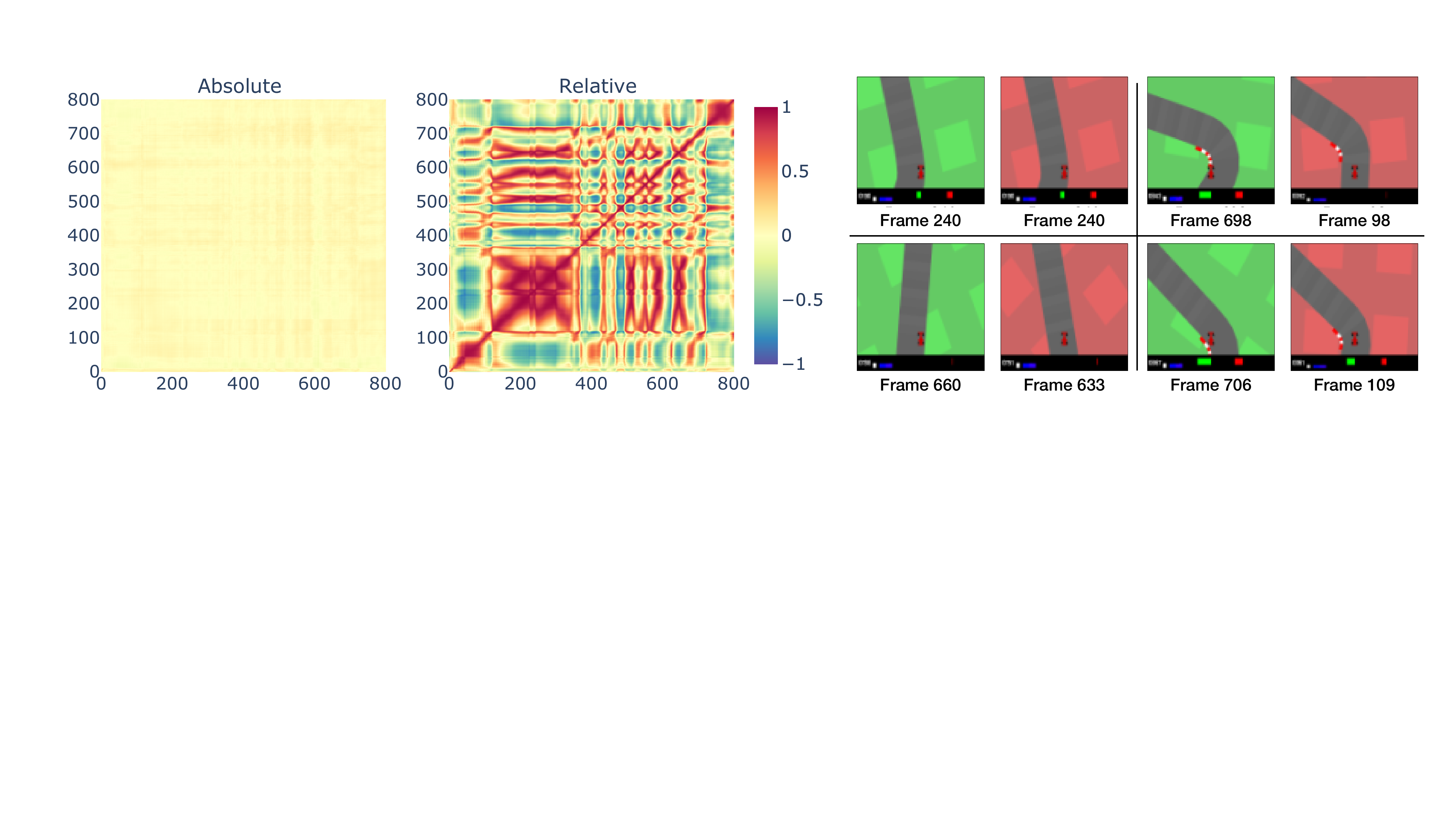}
        \caption{}
        \label{fig:frames-sim}
    \end{subfigure}
    \caption{(a) Comparison between absolute (left) and relative (right) representations produced by the same model. Rows and columns show the cosine similarity between the latent spaces coming from frames of the CarRacing environment with different visual variations (i.e., green and red grass color). Relative representations let similarities emerge not only along the diagonal, where frames are aligned, but also off-diagonal, highlighting similarities between different parts of the track.
    (b) We report qualitative examples by visualizing frame pairs associated to high similarity regions in (a) (denoted by the frame number). Each pair is semantically similar, even though not in direct correspondence.}
    \label{fig:matrix-frames-sim}
\end{figure}

\subsection{Latent space analysis}\label{sec:exp-analysis}

We provide a qualitative analysis of relative representations, showing how the projected spaces of aligned frames with different visual variations exhibit similar latent representations.

We analyze the latent spaces produced by the encoders of two different models trained in the CarRacing environment with different grass colors and seeds. 
To compare the similarity of the latent spaces of frames with two different backgrounds, we collect observations for green and red backgrounds in a similar manner to what we explained in \Cref{sec:datacollection} to collect the anchors, to ensure that the frames are aligned, by playing the same sequence of actions in both environments.
In \Cref{fig:matrix-sim}, we report the pairwise cosine similarities of the first $\sim 800$ frames between the two latent spaces. Therefore, the diagonal shows the similarity between two perfectly aligned frames where the only difference is the grass color. 
As anticipated, the absolute similarities are consistently low, even for the frames along the diagonal. This is expected because the two spaces are not directly comparable. However, when we unify the spaces using relative representations, we observe a strong similarity on the diagonal, along with some similarities off-diagonal. In \Cref{fig:frames-sim}, we present the frames associated with high similarity points in the relative space. Although these points are off-diagonal and not in direct correspondence, they are semantically similar. This shows the effectiveness of relative representations in capturing semantic relationships across different spaces.

In summary, this analysis demonstrates that different policies trained in different contexts exhibit emerging similarities in their latent representations.

\Cref{fig:abs-relrepr-architecture} compares the standard (“absolute”) encoder outputs (left) and the R3L encoder outputs (right) for two different data sets. In the left plot, visual observations corresponding to two different visual variations are embedded into clearly separated regions of the latent space. By contrast, the right plot shows how relative representations brings these data points into much closer alignment, resulting in a near one‐to‐one mapping. This indicates that the relative approach successfully reduces or ignores the differences between the two data sets, focusing on the image semantics and thus producing more consistent and comparable embeddings.




\subsection{End-to-end performance}\label{sec:exp-end-to-end-performance}

\begin{figure*}[t!]
    \centering
    \begin{subfigure}[b]{0.3\textwidth}
        \centering
        \includegraphics[width=\textwidth]{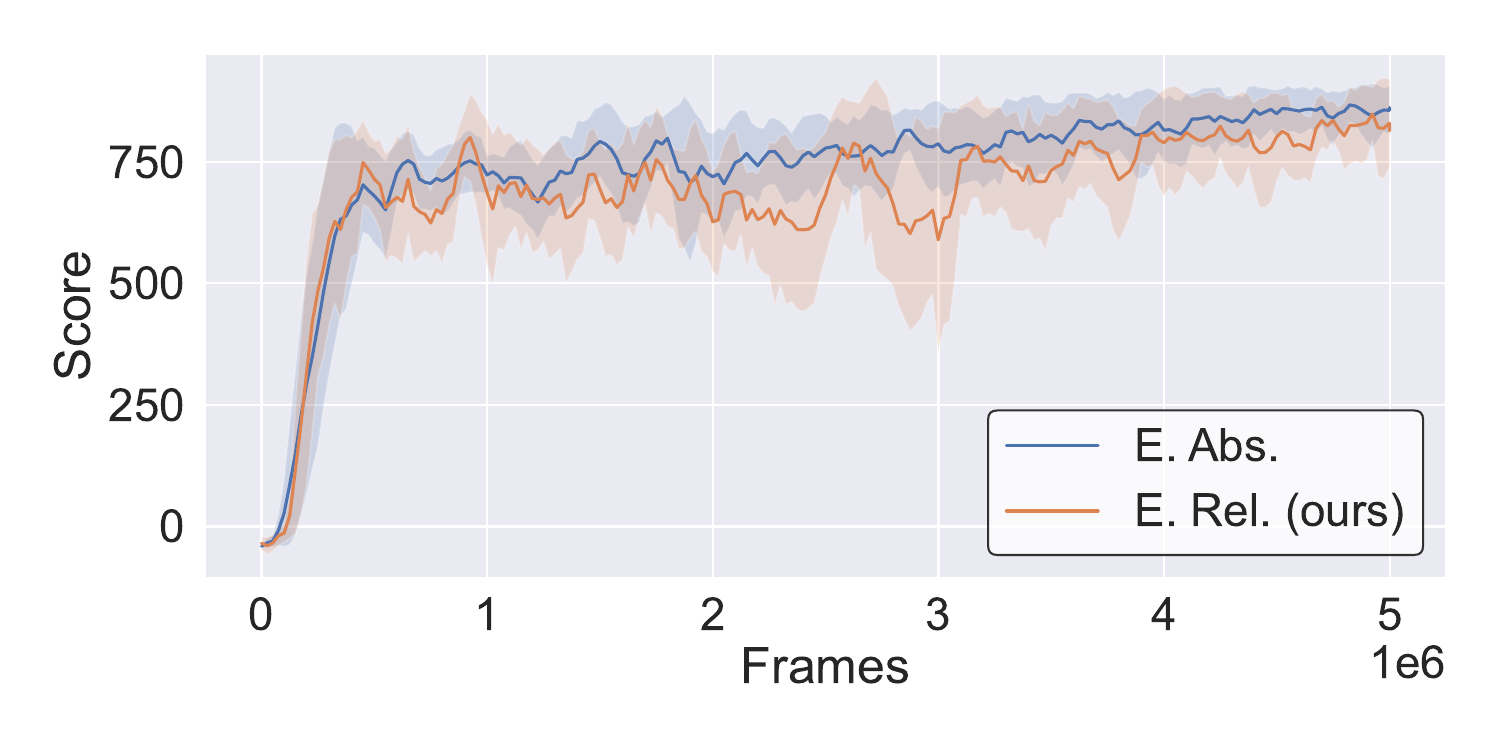}
        \caption{CarRacing (standard)}
        \label{subfig:training-carracing-standard}
    \end{subfigure}
    \hfill
    \begin{subfigure}[b]{0.3\textwidth}
        \centering
        \includegraphics[width=\textwidth]{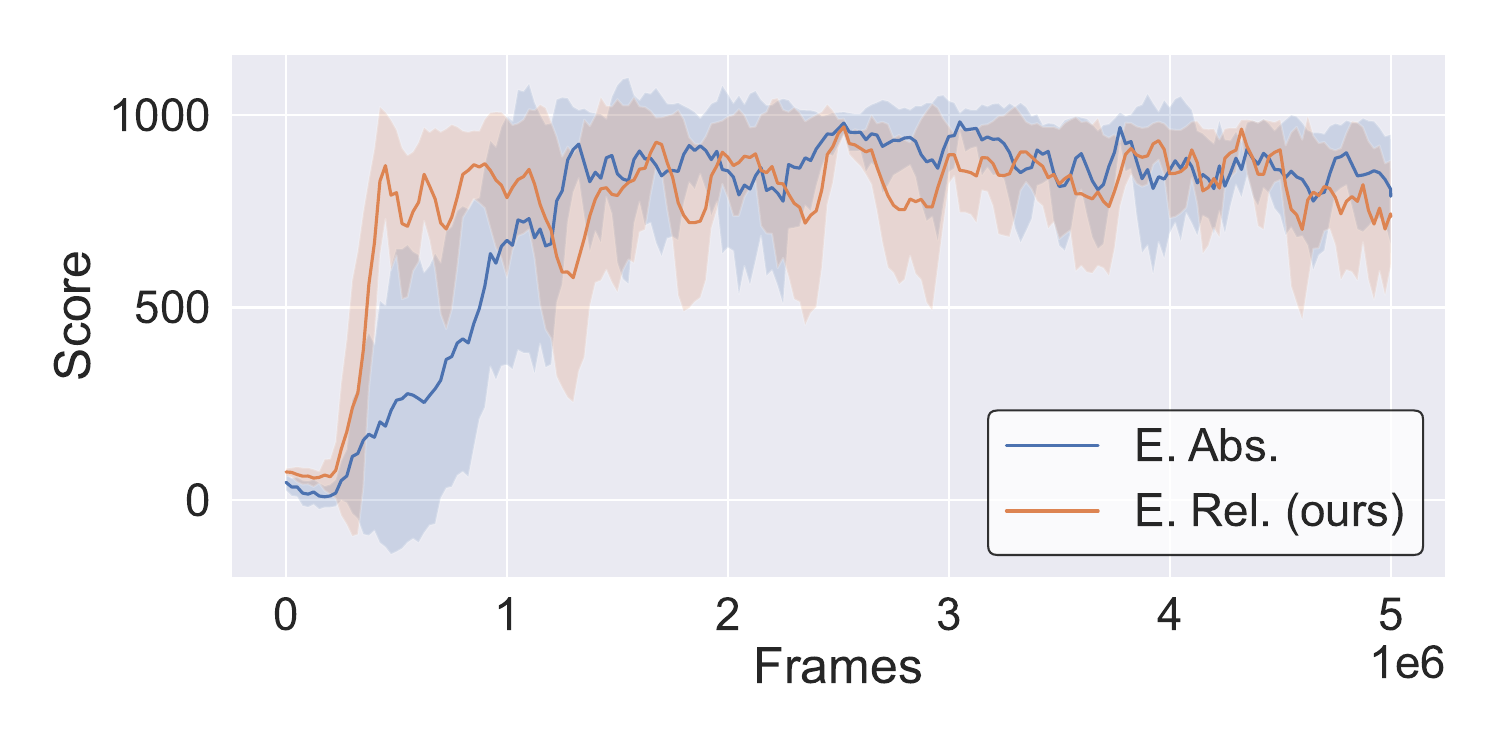}
        \caption{CarRacing (camera far)}
        \label{subfig:training-carracing-camera_far}
    \end{subfigure}
    \hfill
    \begin{subfigure}[b]{0.3\textwidth}
        \centering
        \includegraphics[width=\textwidth]{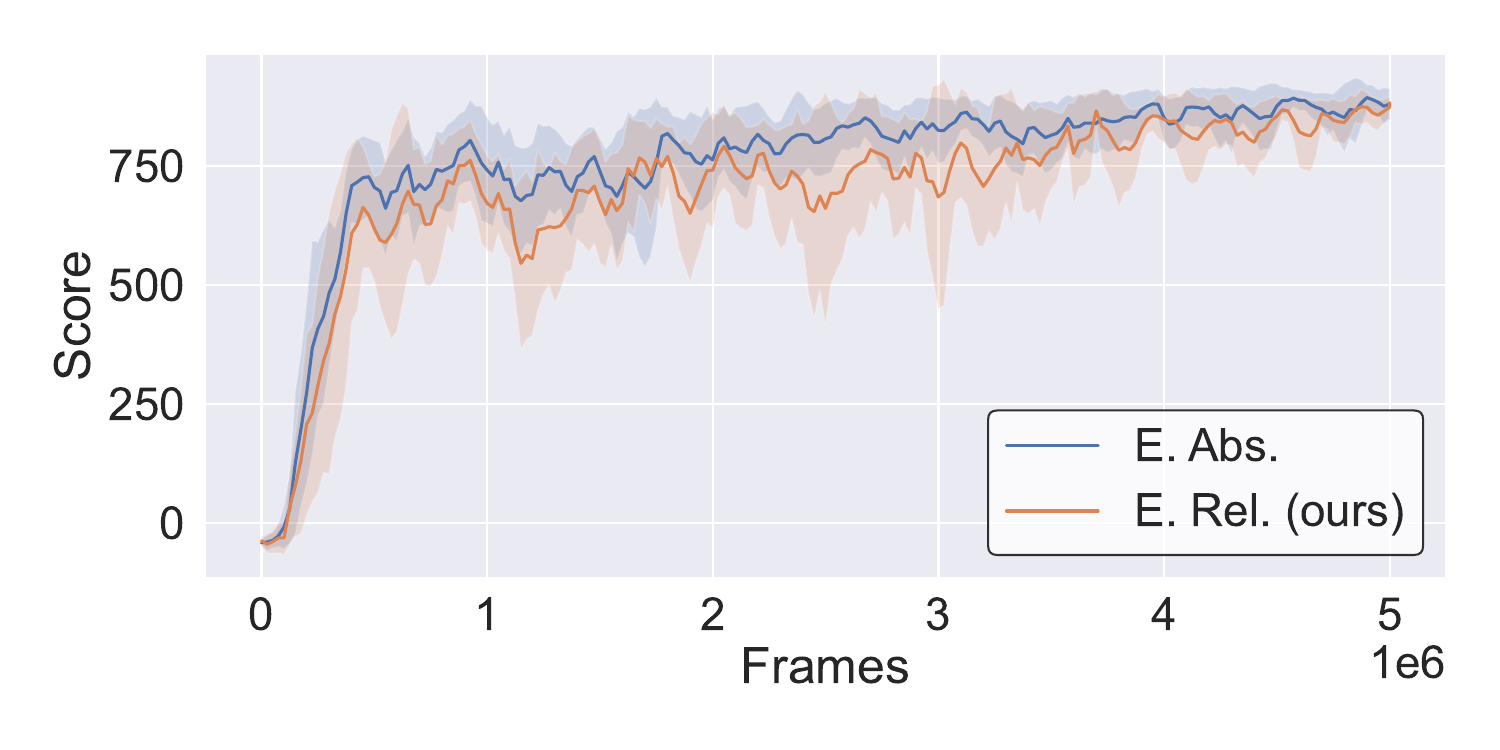}
        \caption{CarRacing (slow)}
        \label{subfig:training-carracing-slow}
    \end{subfigure}
    
    \vspace{5mm} 
    
    \begin{subfigure}[b]{0.3\textwidth}
        \centering
        \includegraphics[width=\textwidth]{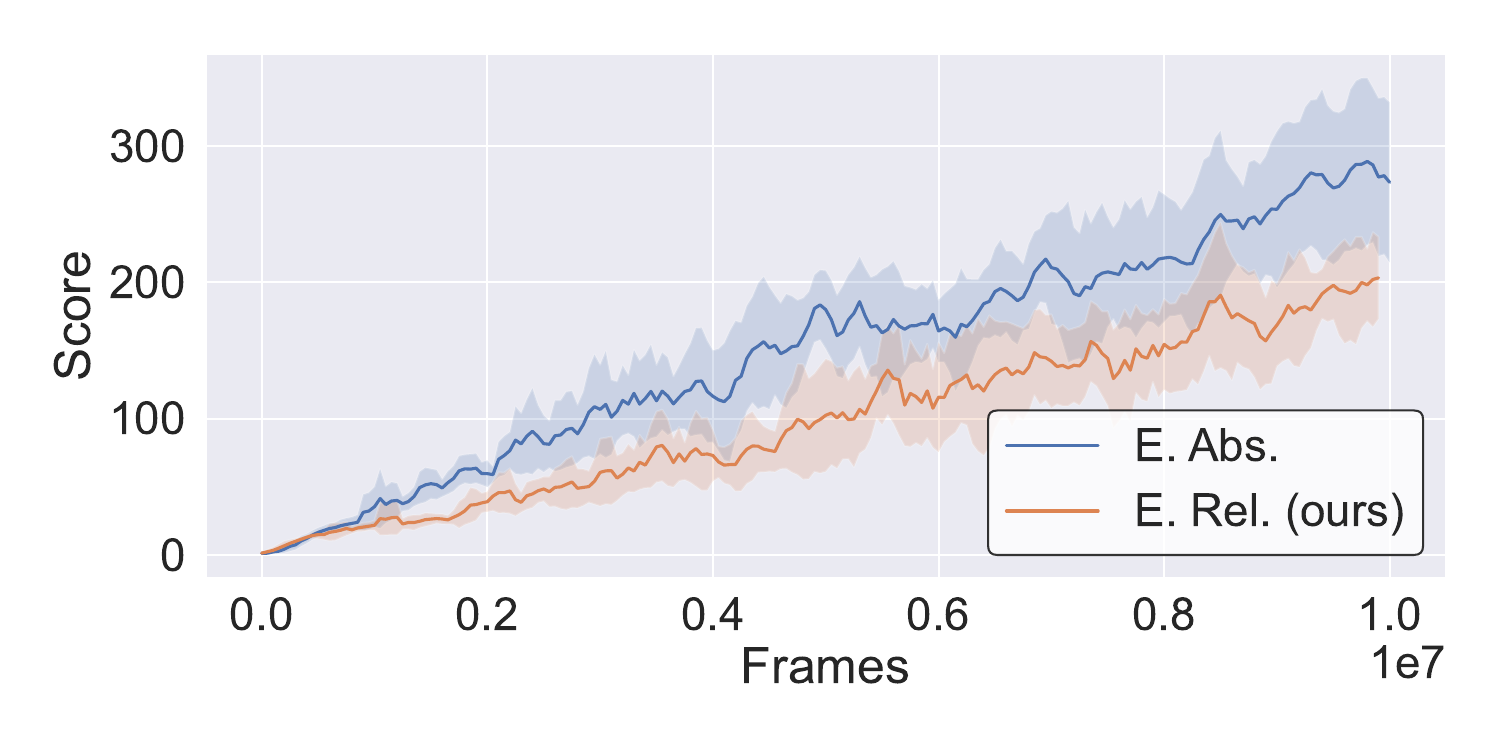}
        \caption{Breakout}
        \label{subfig:training-atari-breakout}
    \end{subfigure}
    \hfill
    \begin{subfigure}[b]{0.3\textwidth}
        \centering
        \includegraphics[width=\textwidth]{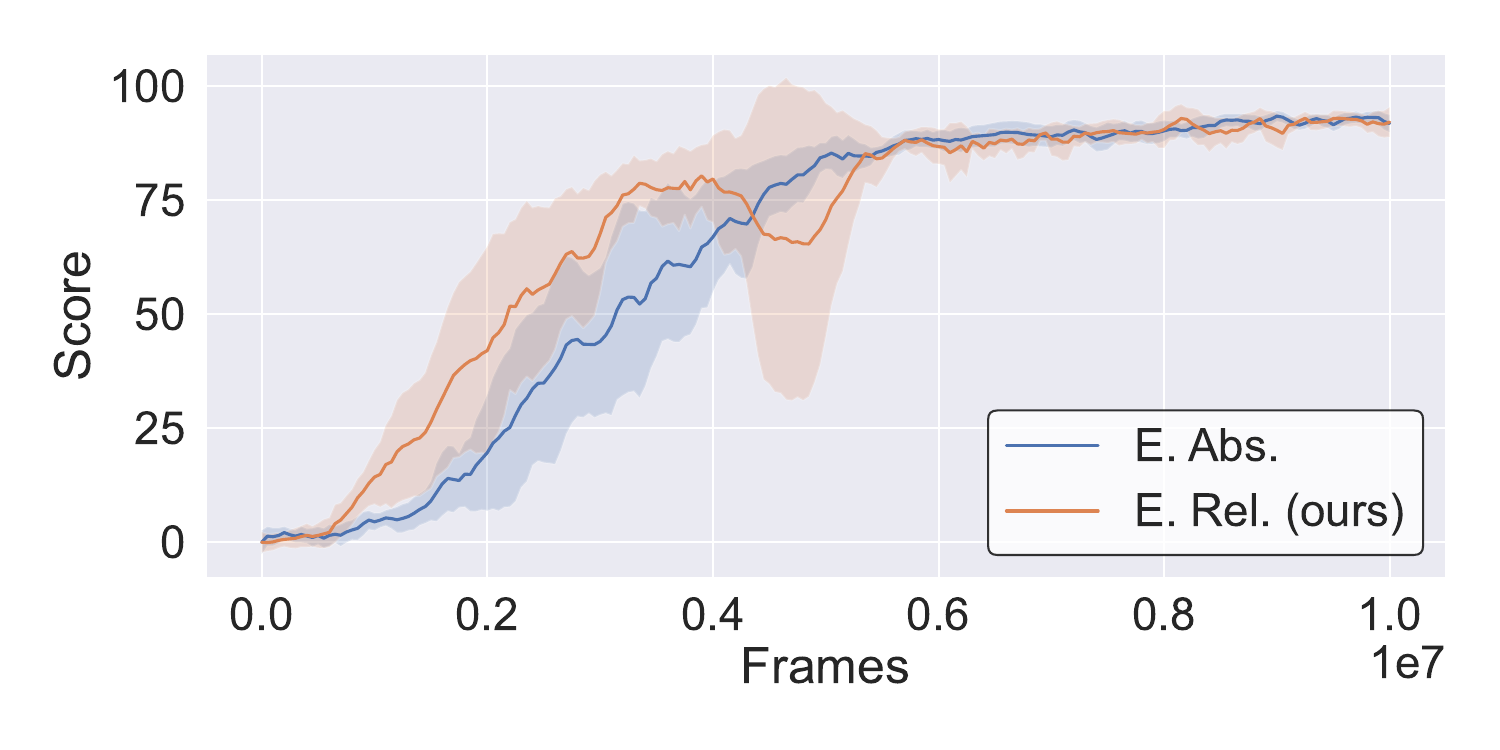}
        \caption{Boxing}
        \label{subfig:training-atari-boxing}
    \end{subfigure}
    \hfill
    \begin{subfigure}[b]{0.3\textwidth}
        \centering
        \includegraphics[width=\textwidth]{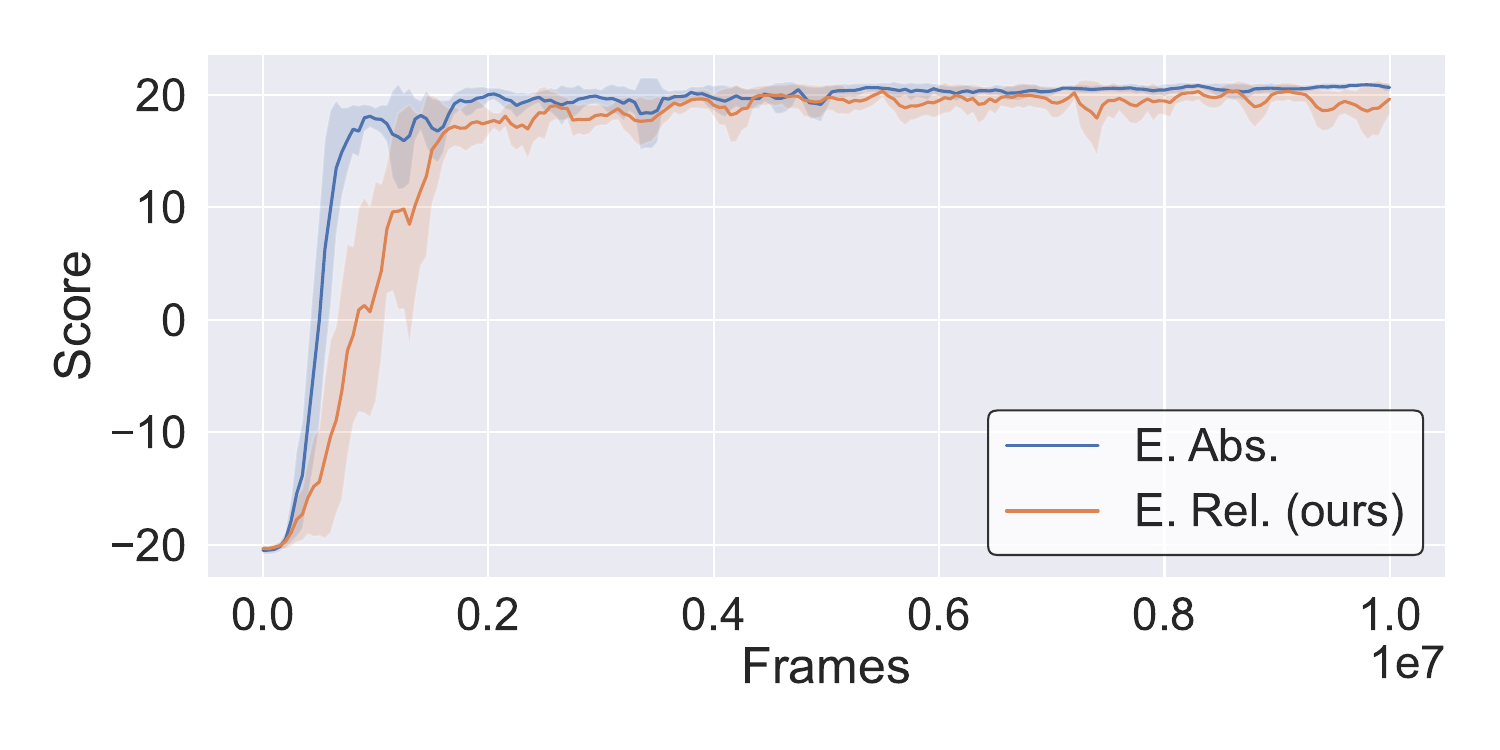}
        \caption{Pong}
        \label{subfig:training-atari-pong}
    \end{subfigure}
    
    \caption{
        \textbf{Comparison of Abs and R3L Training Curves Across Environments.}
        \textbf{Top Row:} Training curves for three variations of the CarRacing environment, demonstrating that both Abs and R3L methods exhibit similar convergence tendencies, indicating that relative encoding does not cause training instability.
        \textbf{Bottom Row:} Training curves for three Atari games (Breakout, Boxing, and Pong), further supporting that both methods maintain stable and comparable performance across different gaming environments.
    }
    \label{fig:rel-carracing-training_comparison}
\end{figure*}

\Cref{fig:rel-carracing-training_comparison} shows the training curves for CarRacing and Atari variations, comparing \emph{Abs.} to \emph{R3L} (\textbf{ours}) under different conditions. The curves are generated using evaluation scores obtained during training, averaged over four different seeds. Solid lines represent the mean values, and shaded areas indicate the standard deviation. The training stability of R3L is comparable to that of standard (absolute) training.
Furthermore, \Cref{tab:relative-carracing-no_stitching} shows end-to-end evaluation performance of the CarRacing models (meaning that models are used as trained, no stitching performed).
This table also includes \emph{Multicolor}, a reference baseline of a model trained on green, red and yellow background. Results demonstrate that the end-to-end performance of agents is generally comparable to those of absolute models, except for the model trained on the slow task. Moreover, \emph{Multicolor} models are unable to generalize to new background colors.

These results show that R3L renders feasible the implementation of relative representations in the context of reinforcement learning rendering the 
evaluation performance generally comparable to those of the end-to-end models. Refer to \Cref{appendix:end-to-end} for end-to-end tests for some games in the Atari game suite.

\begin{table}[h]
\caption{Mean scores for models tested as they were trained, without performing stitching. Models trained with R3L have comparable performance, with small performance loss on average. Scores are computed over four training seeds and, for each combination, over ten distinct tracks}
\label{tab:relative-carracing-no_stitching}
\centering
\begin{subtable}{\columnwidth}
    \resizebox{\linewidth}{!}{%
    \begin{tabular}{lcccc}
    \toprule
    & \texttt{green} & \texttt{red} & \texttt{blue} & \texttt{far (green)} \\
    \midrule
    \textit{E. Abs} & 829 $\pm$ 54 & 854 $\pm$ 26 & 852 $\pm$ 48 & 872 $\pm$ 35 \\
    \textit{E. R3L} (ours) & 832 $\pm$ 54 & 797 $\pm$ 86 & 811 $\pm$ 21 & 820 $\pm$ 22 \\
    \hline
    \textit{E. multicolor} & 839 $\pm$ 127 & 900 $\pm$ 32 & 158 $\pm$ 106 \\
    \bottomrule
    \end{tabular}
    }
    \caption{Visual variations}
    \label{tab:relative-carracing-task}
\end{subtable}
\hfill
\begin{subtable}{\columnwidth}
    \resizebox{\linewidth}{!}{%
    \fontsize{5.6pt}{5.6pt}\selectfont
    \begin{tabular}{lccc}
    \toprule
    & \texttt{slow} & \texttt{scrambled} & \texttt{no idle} \\
    \midrule
    \textit{E. Abs} & 996 $\pm$ 6 & 879 $\pm$ 42 & 889 $\pm$ 19 \\
    \textit{E. R3L} (ours) & 624 $\pm$ 125 & 874 $\pm$ 20 & 862 $\pm$ 69 \\
    \bottomrule
    \end{tabular}
    }
    \caption{Task variations}
    \label{tab:relative-carracing-task}
\end{subtable}
\end{table}

\subsection{Zero-shot stitching}\label{sec:zero-shot-stitching}
The advantage of R3L become most apparent when creating new agents by composing encoders and controllers in a zero-shot fashion, which allows these new agents to operate in environments they have never encountered during training. Indeed, as shown in \Cref{sec:exp-analysis}, models trained with different seeds or under different settings develop distinct latent representations, making it impossible to naively stitch together independently trained encoders and controllers.

\Cref{tab:carracing-stitching_performance} presents the zero-shot stitching performance between encoders and controllers across seed, visual, and task variations. Each component is trained with its own seed. When \textbf{Encoder} and \textbf{Controller} variations are the same (e.g., green-green), we only consider the performance of stitching between different seeds. Visual and task variations are analyzed independently; hence, the \textit{Task Variations (green)} columns only consider controllers originally trained on a green background, while they are stitched to encoders trained on different background colors.
Each cell reports the mean score and standard deviation calculated across ten seeds for the track, four encoders, and four controllers, for a total of 160 aggregated scores per cell.
This table too presents the \textit{multicolor} baseline. 
If in the end-to-end setting, this baseline only failed to generalize to a new color, in this case it always fails to perform in zero-shot stitching.
R3L significantly outperforms the baselines. Interestingly, those agents get a strong performance decline with the \textit{slow} task variation. 

In summary, these findings indicate promising results in using R3L for assembling agents capable of operating in novel environment variations. Again, stitching results for the Atari suite can be seen in \Cref{appendix:stitching-atari}.

\begin{table*}[t!]
    \caption{Comparing R3L and absolute in zero-shot stitching performance.
    The original domains for the encoders and the controllers are listed in the rows and columns, respectively.
    R3L outperforms the absolute model in all the tests.}
    \label{tab:carracing-stitching_performance}
    \centering
    \resizebox{0.8\textwidth}{!}{
    \begin{tabular}{ccccccccccc}
    \toprule
    & & & \multicolumn{6}{c}{\textbf{Controller}} & \\
    \cmidrule{4-9}
    & & & \multicolumn{3}{c}{\textbf{Visual Variations} (standard task)} & \multicolumn{3}{c}{\textbf{Task Variations} (green)} \\
    \cmidrule(r){4-6} \cmidrule(l){7-9}
    & & & \texttt{green} & \texttt{red} & \texttt{blue} &  \texttt{slow} & \texttt{scrambled} & \texttt{no idle} \\
    \cmidrule(r){1-6} \cmidrule(l){7-9}
    \multirow{8}{*}{\rotatebox{90}{\textbf{Encoder}}} 
    & \multirow{2}{*}{\rotatebox{90}{\texttt{green}}} 
    & \emph{S. Abs} & 175 $\pm$ 304 & 167 $\pm$ 226 & -4 $\pm$ 79 &  148 $\pm$ 328 & 106 $\pm$ 217 & 213 $\pm$ 201 \\
    & & \emph{\textbf{S. R3L}} & $\mathbf{781}$ $\pm$ 108 & $\mathbf{787}$ $\pm$ 62 & $\mathbf{794}$ $\pm$ 61 & $\mathbf{268}$ $\pm$ 14 & $\mathbf{781}$ $\pm$ 126 & $\mathbf{824}$ $\pm$ 82 \\[1.5ex]
    & \multirow{2}{*}{\rotatebox{90}{\texttt{red}}} 
    & \emph{S. Abs} & 157 $\pm$ 248 & 43 $\pm$ 205 & 22 $\pm$ 112  & 83 $\pm$ 191 & 138 $\pm$ 244 & 252 $\pm$ 228 \\
    & & \emph{\textbf{S. R3L}} & $\mathbf{810}$ $\pm$ 52 & $\mathbf{776}$ $\pm$ 92 & $\mathbf{803}$ $\pm$ 58 & $\mathbf{476}$ $\pm$ 430 & $\mathbf{790}$ $\pm$ 72 & $\mathbf{817}$ $\pm$ 69 \\[1.5ex]
    & \multirow{2}{*}{\rotatebox{90}{\texttt{blue}}} 
    & \emph{S. Abs} & 137 $\pm$ 225 & 130 $\pm$ 274 & 11 $\pm$ 122 & 95 $\pm$ 128 & 138 $\pm$ 224 & 144 $\pm$ 206 \\
    & & \emph{\textbf{S. R3L}} & $\mathbf{791}$ $\pm$ 64 & $\mathbf{793}$ $\pm$ 40 & $\mathbf{792}$ $\pm$ 48  & $\mathbf{564}$ $\pm$ 440 & $\mathbf{804}$ $\pm$ 41 & $\mathbf{828}$ $\pm$ 50 \\[1.5ex]
    & \multirow{2}{*}{\rotatebox{90}{\texttt{far}}} 
    & \emph{S. Abs} & 152 $\pm$ 204 & 65 $\pm$ 180 & 2 $\pm$ 152 & -49 $\pm$ 9 & 351 $\pm$ 97 & 349 $\pm$ 66 \\
    & & \emph{\textbf{S. R3L}} & $\mathbf{527}$ $\pm$ 142 & $\mathbf{605}$ $\pm$ 118 & $\mathbf{592}$ $\pm$ 86 & $\mathbf{303}$ $\pm$ 100 & $\mathbf{594}$ $\pm$ 39 & $\mathbf{673}$ $\pm$ 91 \\[1.5ex]
    \hline
    & & \emph{S. multicolor} & -67 $\pm$ 41 & -59 $\pm$ 54 & 13 $\pm$ 161 &  $\pm$  & - $\pm$ - & - $\pm$ - \\
    \bottomrule
    \end{tabular}
    }
\end{table*}

\subsubsection{Computational Advantage}
The proposed methods offer a significant computational advantage by reducing the training time required to develop new policies. Indeed, they enable the assembly of new policies from existing ones, making it possible to adapt to novel environments more efficiently.

By reusing policy components, we can create new policies without starting training from scratch. \Cref{tab:carracing-time-saved} illustrates the amount of time saved through zero-shot stitching for the CarRacing models. This table shows the training time required for agents across all visual-task combinations, as previously detailed in \Cref{tab:carracing-stitching_performance}. Normally, training models for every visual-task combination would require 52 hours. However, our approach significantly reduces this time, needing only the highlighted cells in light blue, representing a fraction of the total training time.
Indeed, it is sufficient to have at least one encoder and one controller for each variation. This enables the creation of all other agents, saving 88 hours of training. Importantly, this time-saving benefit scales quadratically with the number of visual variations and tasks considered, providing substantial efficiency gains as the complexity of the environment increases.

In summary, latent communication significantly reduces training time in RL by allowing the reuse of policy components to assemble novel policies without the need to train from scratch.

\begin{table}[h]
\caption{Table of training times. We only need to train combinations for the cells highlighted in blue and then perform zero-shot stitching to assemble all the  other agents (\emph{totaling 13 hrs}). Normally, it would be required to train agents for each domain-task combination (\emph{totaling 52 hrs}). Visual variations \emph{V1}: green, \emph{V2}: red, \emph{V3}: blue, \emph{V4}: far (green) and task variations \emph{T1}: standard, \emph{T2}: slow, \emph{T3}: no idle, \emph{T4}: scrambled.}
\label{tab:carracing-time-saved}
\centering
\begin{tabular}{ccccc}
\toprule
& V1 & V2 & V3 & V4 \\ 
\midrule
T1 & \cellcolor{cellblue}3h & 3h & 3h & 3h \\ 
T2 & 4h & \cellcolor{cellblue}4h & 4h & 4h \\ 
T3 & 3h & 3h & \cellcolor{cellblue}3h & 3h \\ 
T4 & 3h & 3h & 3h & \cellcolor{cellblue}3h \\ 
\bottomrule
\end{tabular}
\end{table}

\subsection{Ablation test}\label{sec:finding-ema}
\begin{figure}[!t]
        \centering        \includegraphics[width=0.45\textwidth]{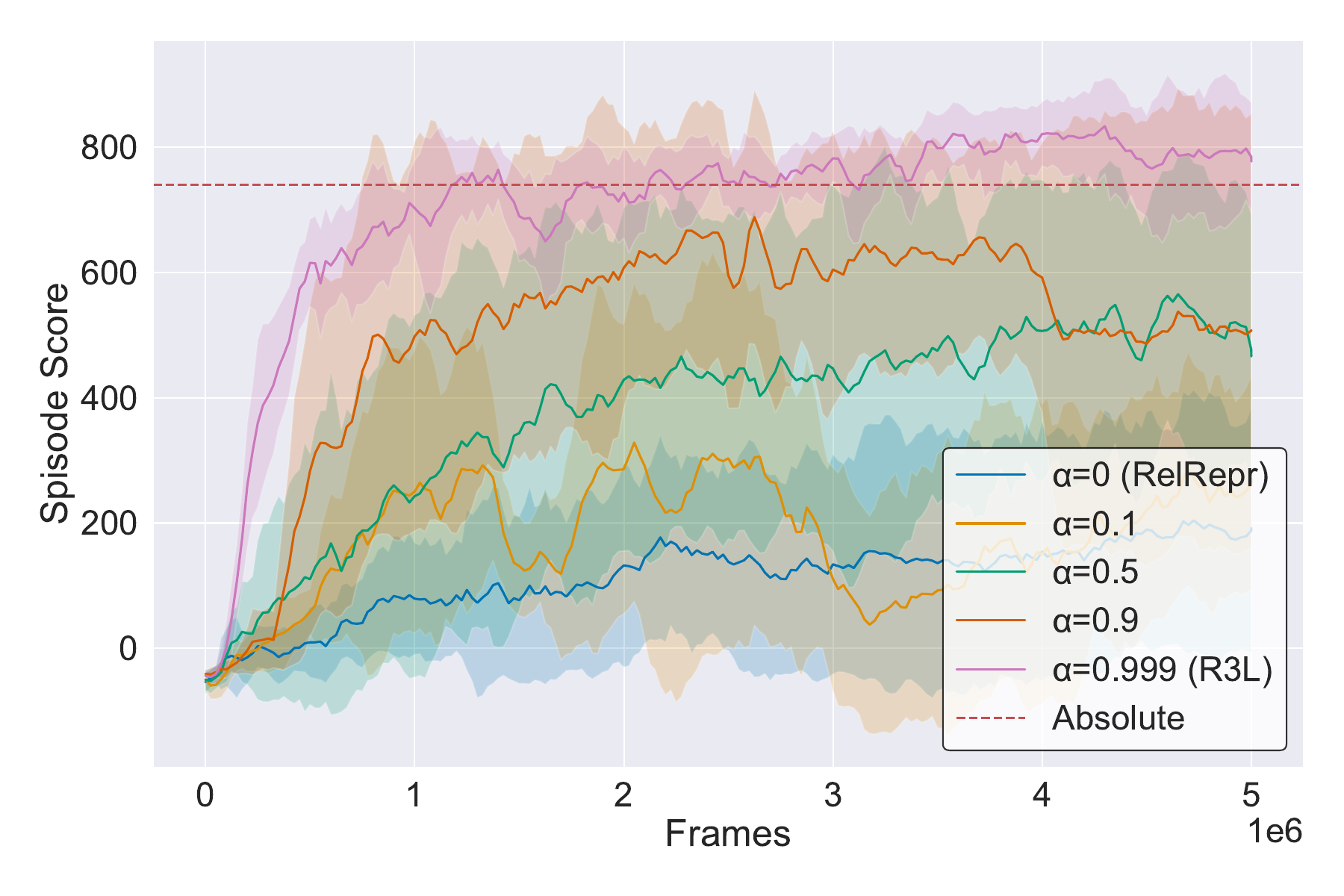}
        \caption{Comparison of evaluation scores over training frames using different values of the exponential moving average coefficient ($\alpha$). Solid lines represent mean evaluation scores, shaded regions indicate standard deviations, and the dashed red line denotes the absolute evaluation score.}
        \label{subfig:training-carracing-ema}
\end{figure}

\paragraph{Finding the best $\alpha$ for EMA}
We trained models on CarRacing with various $\alpha$ to find the best value to use with exponential moving average function. \Cref{subfig:training-carracing-ema} shows a comparison with different values.

\section{Conclusion and limitations}

\paragraph{Limitations and Future Works} 
We believe our work opens to several compelling research directions. We restricted our analysis to controlled environments, therefore extending our methodology to more complex and real environments would be of paramount importance to provide further insights into its scalability. Similarly, all our agents are trained from scratch on specific tasks for consistency. Relying instead on large pre-trained vision models and stitching to different controllers would provide even further computational savings and flexibility.

\paragraph{Conclusions}
With R3L, we showed how to adapt relative representations in the context of visual RL, demonstrating the necessity of using the exponential moving average to stabilize training. We performed an analysis of the latent spaces, showing how relative representations allow to generate universal embeddings, and that those can be used to combine encoders and controllers of models trained via R3L, in a zero-shot fashion, to create new agents that can act on visual-task pairs never seen at training time.
Our approach provides a paradigm that could make RL more accessible, thanks to the possibility of reducing computational costs via zero-shot stitching.

\section*{Impact Statement}
R3L enables the creation of new reinforcement learning agents by combining encoders and controllers across visual and task variations without fine-tuning, paving the way for more flexible, reusable, and scalable AI systems.
\bibliography{bibliography}

\begin{thebibliography}{44}
\providecommand{\natexlab}[1]{#1}
\providecommand{\url}[1]{\texttt{#1}}
\expandafter\ifx\csname urlstyle\endcsname\relax
  \providecommand{\doi}[1]{doi: #1}\else
  \providecommand{\doi}{doi: \begingroup \urlstyle{rm}\Url}\fi

\bibitem[Bansal et~al.(2021)Bansal, Nakkiran, and Barak]{Bansal2021-oj}
Bansal, Y., Nakkiran, P., and Barak, B.
\newblock Revisiting model stitching to compare neural representations.
\newblock In Ranzato, M., Beygelzimer, A., Dauphin, Y.~N., Liang, P., and Vaughan, J.~W. (eds.), \emph{Advances in Neural Information Processing Systems 34: Annual Conference on Neural Information Processing Systems 2021, NeurIPS 2021, December 6-14, 2021, virtual}, pp.\  225--236, 2021.

\bibitem[Barreto et~al.(2017)Barreto, Dabney, Munos, Hunt, Schaul, van Hasselt, and Silver]{barreto2017successor}
Barreto, A., Dabney, W., Munos, R., Hunt, J.~J., Schaul, T., van Hasselt, H.~P., and Silver, D.
\newblock Successor features for transfer in reinforcement learning.
\newblock \emph{Advances in neural information processing systems}, 30, 2017.

\bibitem[Cannistraci et~al.(2024)Cannistraci, Moschella, Fumero, Maiorca, and Rodol\`a]{cannistraci2023bricks}
Cannistraci, I., Moschella, L., Fumero, M., Maiorca, V., and Rodol\`a, E.
\newblock From bricks to bridges: Product of invariances to enhance latent space communication.
\newblock In \emph{Proc. ICLR}, 2024.

\bibitem[Csiszarik et~al.(2021)Csiszarik, Korosi-Szabo, Matszangosz, Papp, and Varga]{Csiszarik2021-yi}
Csiszarik, A., Korosi-Szabo, P., Matszangosz, A.~K., Papp, G., and Varga, D.
\newblock Similarity and matching of neural network representations.
\newblock \emph{ArXiv preprint}, abs/2110.14633, 2021.

\bibitem[Devin et~al.(2017)Devin, Gupta, Darrell, Abbeel, and Levine]{devin2017learning}
Devin, C., Gupta, A., Darrell, T., Abbeel, P., and Levine, S.
\newblock Learning modular neural network policies for multi-task and multi-robot transfer.
\newblock In \emph{2017 IEEE international conference on robotics and automation (ICRA)}, pp.\  2169--2176. IEEE, 2017.

\bibitem[Fern{\'a}ndez \& Veloso(2006)Fern{\'a}ndez and Veloso]{fernandez2006probabilistic}
Fern{\'a}ndez, F. and Veloso, M.
\newblock Probabilistic policy reuse in a reinforcement learning agent.
\newblock In \emph{Proceedings of the fifth international joint conference on Autonomous agents and multiagent systems}, pp.\  720--727, 2006.

\bibitem[Finn et~al.(2017)Finn, Abbeel, and Levine]{finn2017model}
Finn, C., Abbeel, P., and Levine, S.
\newblock Model-agnostic meta-learning for fast adaptation of deep networks.
\newblock In \emph{International conference on machine learning}, pp.\  1126--1135. PMLR, 2017.

\bibitem[Goyal et~al.(2019)Goyal, Lamb, Hoffmann, Sodhani, Levine, Bengio, and Sch{\"o}lkopf]{goyal2019recurrent}
Goyal, A., Lamb, A., Hoffmann, J., Sodhani, S., Levine, S., Bengio, Y., and Sch{\"o}lkopf, B.
\newblock Recurrent independent mechanisms.
\newblock \emph{arXiv preprint arXiv:1909.10893}, 2019.

\bibitem[Gygli et~al.(2021)Gygli, Uijlings, and Ferrari]{Gygli2021-qb}
Gygli, M., Uijlings, J., and Ferrari, V.
\newblock Towards reusable network components by learning compatible representations.
\newblock \emph{AAAI}, 35\penalty0 (9):\penalty0 7620--7629, 2021.

\bibitem[Hansen \& Wang(2021)Hansen and Wang]{hansen2021softda}
Hansen, N. and Wang, X.
\newblock Generalization in reinforcement learning by soft data augmentation.
\newblock In \emph{International Conference on Robotics and Automation}, 2021.

\bibitem[Hansen et~al.(2020)Hansen, Jangir, Sun, Aleny{\`a}, Abbeel, Efros, Pinto, and Wang]{hansen2020self}
Hansen, N., Jangir, R., Sun, Y., Aleny{\`a}, G., Abbeel, P., Efros, A.~A., Pinto, L., and Wang, X.
\newblock Self-supervised policy adaptation during deployment.
\newblock \emph{arXiv preprint arXiv:2007.04309}, 2020.

\bibitem[Hansen et~al.(2021)Hansen, Su, and Wang]{hansen2021stabilizing}
Hansen, N., Su, H., and Wang, X.
\newblock Stabilizing deep q-learning with convnets and vision transformers under data augmentation.
\newblock In \emph{Conference on Neural Information Processing Systems}, 2021.

\bibitem[Huang et~al.(2022)Huang, Dossa, Ye, Braga, Chakraborty, Mehta, and Araújo]{huang2022cleanrl}
Huang, S., Dossa, R. F.~J., Ye, C., Braga, J., Chakraborty, D., Mehta, K., and Araújo, J.~G.
\newblock Cleanrl: High-quality single-file implementations of deep reinforcement learning algorithms.
\newblock \emph{Journal of Machine Learning Research}, 23\penalty0 (274):\penalty0 1--18, 2022.
\newblock URL \url{http://jmlr.org/papers/v23/21-1342.html}.

\bibitem[Jian et~al.(2021)Jian, Yang, Guo, Liu, and Sun]{jian2021adversarial}
Jian, P., Yang, C., Guo, D., Liu, H., and Sun, F.
\newblock Adversarial skill learning for robust manipulation.
\newblock In \emph{2021 IEEE International Conference on Robotics and Automation (ICRA)}, pp.\  2555--2561. IEEE, 2021.

\bibitem[Jian et~al.(2023)Jian, Lee, Bell, Zavlanos, and Chen]{jian2023policy}
Jian, P., Lee, E., Bell, Z., Zavlanos, M.~M., and Chen, B.
\newblock Policy stitching: Learning transferable robot policies.
\newblock \emph{arXiv preprint arXiv:2309.13753}, 2023.

\bibitem[Karkus et~al.(2020)Karkus, Mirza, Guez, Jaegle, Lillicrap, Buesing, Heess, and Weber]{karkus2020beyond}
Karkus, P., Mirza, M., Guez, A., Jaegle, A., Lillicrap, T., Buesing, L., Heess, N., and Weber, T.
\newblock Beyond tabula-rasa: a modular reinforcement learning approach for physically embedded 3d sokoban.
\newblock \emph{arXiv preprint arXiv:2010.01298}, 2020.

\bibitem[Killian et~al.(2017)Killian, Daulton, Konidaris, and Doshi-Velez]{killian2017robust}
Killian, T.~W., Daulton, S., Konidaris, G., and Doshi-Velez, F.
\newblock Robust and efficient transfer learning with hidden parameter markov decision processes.
\newblock \emph{Advances in neural information processing systems}, 30, 2017.

\bibitem[Klimov(2016)]{klimov2016carracing}
Klimov, O.
\newblock Carracing-v0.
\newblock \emph{URL https://gym. openai. com/envs/CarRacing-v0}, 2016.

\bibitem[Lenc \& Vedaldi(2015)Lenc and Vedaldi]{Lenc2014-gy}
Lenc, K. and Vedaldi, A.
\newblock Understanding image representations by measuring their equivariance and equivalence.
\newblock In \emph{{IEEE} Conference on Computer Vision and Pattern Recognition, {CVPR} 2015, Boston, MA, USA, June 7-12, 2015}, pp.\  991--999. {IEEE} Computer Society, 2015.
\newblock \doi{10.1109/CVPR.2015.7298701}.

\bibitem[Lillicrap(2015)]{lillicrap2015continuous}
Lillicrap, T.
\newblock Continuous control with deep reinforcement learning.
\newblock \emph{arXiv preprint arXiv:1509.02971}, 2015.

\bibitem[Liu et~al.(2021)Liu, Zhang, Shen, and Zavlanos]{liu2021learning}
Liu, C., Zhang, Y., Shen, Y., and Zavlanos, M.~M.
\newblock Learning without knowing: Unobserved context in continuous transfer reinforcement learning.
\newblock In \emph{Learning for Dynamics and Control}, pp.\  791--802. PMLR, 2021.

\bibitem[Maiorca et~al.(2023)Maiorca, Moschella, Norelli, Fumero, Locatello, and Rodol{\`a}]{maiorca2023latent}
Maiorca, V., Moschella, L., Norelli, A., Fumero, M., Locatello, F., and Rodol{\`a}, E.
\newblock Latent space translation via semantic alignment.
\newblock In \emph{Thirty-seventh Conference on Neural Information Processing Systems}, 2023.
\newblock URL \url{https://openreview.net/forum?id=pBa70rGHlr}.

\bibitem[Mendez et~al.(2022)Mendez, van Seijen, and Eaton]{mendez2022modular}
Mendez, J.~A., van Seijen, H., and Eaton, E.
\newblock Modular lifelong reinforcement learning via neural composition.
\newblock \emph{arXiv preprint arXiv:2207.00429}, 2022.

\bibitem[Mnih et~al.(2015)Mnih, Kavukcuoglu, Silver, Rusu, Veness, Bellemare, Graves, Riedmiller, Fidjeland, Ostrovski, et~al.]{mnih2015human}
Mnih, V., Kavukcuoglu, K., Silver, D., Rusu, A.~A., Veness, J., Bellemare, M.~G., Graves, A., Riedmiller, M., Fidjeland, A.~K., Ostrovski, G., et~al.
\newblock Human-level control through deep reinforcement learning.
\newblock \emph{nature}, 518\penalty0 (7540):\penalty0 529--533, 2015.

\bibitem[Mohanty et~al.(2021)Mohanty, Poonganam, Gaidon, Kolobov, Wulfe, Chakraborty, {\v{S}}emetulskis, Schapke, Kubilius, Pa{\v{s}}ukonis, et~al.]{mohanty2021measuring}
Mohanty, S., Poonganam, J., Gaidon, A., Kolobov, A., Wulfe, B., Chakraborty, D., {\v{S}}emetulskis, G., Schapke, J., Kubilius, J., Pa{\v{s}}ukonis, J., et~al.
\newblock Measuring sample efficiency and generalization in reinforcement learning benchmarks: Neurips 2020 procgen benchmark.
\newblock \emph{arXiv preprint arXiv:2103.15332}, 2021.

\bibitem[Moschella et~al.(2023)Moschella, Maiorca, Fumero, Norelli, Locatello, and Rodol{\`a}]{Moschella2022-yf}
Moschella, L., Maiorca, V., Fumero, M., Norelli, A., Locatello, F., and Rodol{\`a}, E.
\newblock Relative representations enable zero-shot latent space communication.
\newblock In \emph{International Conference on Learning Representations}, 2023.
\newblock URL \url{https://openreview.net/forum?id=SrC-nwieGJ}.

\bibitem[Norelli et~al.(2023)Norelli, Fumero, Maiorca, Moschella, Rodol\`a, and Locatello]{norelli2022b}
Norelli, A., Fumero, M., Maiorca, V., Moschella, L., Rodol\`a, E., and Locatello, F.
\newblock {{ASIF}}: {{Coupled Data Turns Unimodal Models}} to {{Multimodal Without Training}}.
\newblock In \emph{Proc. NeurIPS}, 2023.

\bibitem[Russell \& Zimdars(2003)Russell and Zimdars]{russell2003q}
Russell, S.~J. and Zimdars, A.
\newblock Q-decomposition for reinforcement learning agents.
\newblock In \emph{Proceedings of the 20th International Conference on Machine Learning (ICML-03)}, pp.\  656--663, 2003.

\bibitem[Silver et~al.(2016)Silver, Huang, Maddison, Guez, Sifre, Van Den~Driessche, Schrittwieser, Antonoglou, Panneershelvam, Lanctot, et~al.]{silver2016mastering}
Silver, D., Huang, A., Maddison, C.~J., Guez, A., Sifre, L., Van Den~Driessche, G., Schrittwieser, J., Antonoglou, I., Panneershelvam, V., Lanctot, M., et~al.
\newblock Mastering the game of go with deep neural networks and tree search.
\newblock \emph{nature}, 529\penalty0 (7587):\penalty0 484--489, 2016.

\bibitem[Silver et~al.(2017)Silver, Schrittwieser, Simonyan, Antonoglou, Huang, Guez, Hubert, Baker, Lai, Bolton, et~al.]{silver2017mastering}
Silver, D., Schrittwieser, J., Simonyan, K., Antonoglou, I., Huang, A., Guez, A., Hubert, T., Baker, L., Lai, M., Bolton, A., et~al.
\newblock Mastering the game of go without human knowledge.
\newblock \emph{nature}, 550\penalty0 (7676):\penalty0 354--359, 2017.

\bibitem[Simpkins \& Isbell(2019)Simpkins and Isbell]{simpkins2019composable}
Simpkins, C. and Isbell, C.
\newblock Composable modular reinforcement learning.
\newblock In \emph{Proceedings of the AAAI conference on artificial intelligence}, volume~33, pp.\  4975--4982, 2019.

\bibitem[Sutton(2018)]{sutton2018reinforcement}
Sutton, R.~S.
\newblock Reinforcement learning: An introduction.
\newblock \emph{A Bradford Book}, 2018.

\bibitem[Sutton et~al.(2011)Sutton, Modayil, Delp, Degris, Pilarski, White, and Precup]{sutton2011horde}
Sutton, R.~S., Modayil, J., Delp, M., Degris, T., Pilarski, P.~M., White, A., and Precup, D.
\newblock Horde: A scalable real-time architecture for learning knowledge from unsupervised sensorimotor interaction.
\newblock In \emph{The 10th International Conference on Autonomous Agents and Multiagent Systems-Volume 2}, pp.\  761--768, 2011.

\bibitem[Tirinzoni et~al.(2018)Tirinzoni, Rodriguez~Sanchez, and Restelli]{tirinzoni2018transfer}
Tirinzoni, A., Rodriguez~Sanchez, R., and Restelli, M.
\newblock Transfer of value functions via variational methods.
\newblock \emph{Advances in Neural Information Processing Systems}, 31, 2018.

\bibitem[Tobin et~al.(2017)Tobin, Fong, Ray, Schneider, Zaremba, and Abbeel]{tobin2017domain}
Tobin, J., Fong, R., Ray, A., Schneider, J., Zaremba, W., and Abbeel, P.
\newblock Domain randomization for transferring deep neural networks from simulation to the real world.
\newblock In \emph{2017 IEEE/RSJ international conference on intelligent robots and systems (IROS)}, pp.\  23--30. IEEE, 2017.

\bibitem[Vinyals et~al.(2019)Vinyals, Babuschkin, Czarnecki, Mathieu, Dudzik, Chung, Choi, Powell, Ewalds, Georgiev, et~al.]{vinyals2019grandmaster}
Vinyals, O., Babuschkin, I., Czarnecki, W.~M., Mathieu, M., Dudzik, A., Chung, J., Choi, D.~H., Powell, R., Ewalds, T., Georgiev, P., et~al.
\newblock Grandmaster level in starcraft ii using multi-agent reinforcement learning.
\newblock \emph{Nature}, 575\penalty0 (7782):\penalty0 350--354, 2019.

\bibitem[Wolf \& Musolesi(2023)Wolf and Musolesi]{wolf2023augmented}
Wolf, L. and Musolesi, M.
\newblock Augmented modular reinforcement learning based on heterogeneous knowledge.
\newblock \emph{arXiv preprint arXiv:2306.01158}, 2023.

\bibitem[Yaman et~al.(2022)Yaman, Kalinin, Guye, Ginger, and Ziatdinov]{https://doi.org/10.48550/arxiv.2208.03861}
Yaman, M.~Y., Kalinin, S.~V., Guye, K.~N., Ginger, D., and Ziatdinov, M.
\newblock Learning and predicting photonic responses of plasmonic nanoparticle assemblies via dual variational autoencoders.
\newblock \emph{ArXiv preprint}, abs/2208.03861, 2022.
\newblock URL \url{https://arxiv.org/abs/2208.03861}.

\bibitem[Yarats et~al.(2021{\natexlab{a}})Yarats, Fergus, Lazaric, and Pinto]{yarats2021drqv2}
Yarats, D., Fergus, R., Lazaric, A., and Pinto, L.
\newblock Mastering visual continuous control: Improved data-augmented reinforcement learning.
\newblock \emph{arXiv preprint arXiv:2107.09645}, 2021{\natexlab{a}}.

\bibitem[Yarats et~al.(2021{\natexlab{b}})Yarats, Kostrikov, and Fergus]{yarats2021image}
Yarats, D., Kostrikov, I., and Fergus, R.
\newblock Image augmentation is all you need: Regularizing deep reinforcement learning from pixels.
\newblock In \emph{International Conference on Learning Representations}, 2021{\natexlab{b}}.
\newblock URL \url{https://openreview.net/forum?id=GY6-6sTvGaf}.

\bibitem[Yoneda et~al.(2021)Yoneda, Yang, Walter, and Stadie]{yoneda2021invariance}
Yoneda, T., Yang, G., Walter, M.~R., and Stadie, B.
\newblock Invariance through latent alignment.
\newblock \emph{arXiv preprint arXiv:2112.08526}, 2021.

\bibitem[Zhang et~al.(2018)Zhang, Wu, and Pineau]{zhang2018natural}
Zhang, A., Wu, Y., and Pineau, J.
\newblock Natural environment benchmarks for reinforcement learning.
\newblock \emph{arXiv preprint arXiv:1811.06032}, 2018.

\bibitem[Zhang et~al.(2020)Zhang, McAllister, Calandra, Gal, and Levine]{zhang2020learning}
Zhang, A., McAllister, R., Calandra, R., Gal, Y., and Levine, S.
\newblock Learning invariant representations for reinforcement learning without reconstruction.
\newblock \emph{arXiv preprint arXiv:2006.10742}, 2020.

\bibitem[Zhu et~al.(2023)Zhu, Lin, Jain, and Zhou]{zhu2023transfer}
Zhu, Z., Lin, K., Jain, A.~K., and Zhou, J.
\newblock Transfer learning in deep reinforcement learning: A survey.
\newblock \emph{IEEE Transactions on Pattern Analysis and Machine Intelligence}, 2023.

\end{thebibliography}
\bibliographystyle{icml2025}

\newpage
\appendix
\onecolumn

\section{Appendix}
\subsection{Environments}

\paragraph{CarRacing}
We use the discretized version of CarRacing, which has the following Action space: \textbf{0:} left, \textbf{1:} right, \textbf{2:} accelerate, \textbf{3:} brake, \textbf{4:} do nothing.
We use the standard dynamics, with small rewards for passing through checkpoints and a small penalty every step otherwise, -100, and end episode if the car goes out of the boundaries. Normal car density (the "weight" of the car) is 1.0. Zoom $= 2.7$. During our trials, we perform visual (grass color, camera with zoom=1) and behavior variations. The latter are performed via reward constraints (car speed), or by modifying the action space (scrambled actions, smaller action space).

\begin{table}[h]
    \caption{Complete list of variations applied for the CarRacing environment.}
    \centering
    \begin{tabular}{c c c c c}
        \hline
        \textbf{Variation} & \textbf{Type} & \textbf{ep. length} & \textbf{Other Details} \\
        \hline
        \textit{color} & visual & 1000 & colors: green, red, blue\\
        \textit{camera} & visual & 1000 & far view (zoom = 1) \\
        \textit{slow} & task & 3000 & no negative reward per step, $-100$ if speed $> 35$. \\
        \textit{scrambled} & task & 1000 & action space: shuffled \\
        \textit{no idle} & task & 1000 & action space: noop action removed \\
    \end{tabular}
    \label{tab:appendix-carracing-envs}
\end{table}

Models for all visual and task variations are trained with the same set of hyperparameters

\subsection{Atari Games}
We base our environment suite on NaturalEnv \cite{zhang2018natural}, which originally allowed to replace default game background with images or solid colors. Our selection of games is: \emph{Breakout, Boxing, Pong}. For all we select the \textit{NoFrameskip-v4} version, with default action space. Models are trained for 10000000 steps and default training parameters as in \citep{huang2022cleanrl}.


\subsection{Latent communication on the Atari game suite}\label{appendix:atari}
To test the generalizability of our method we also perform stitching tests with the following Atari games: Breakout, Boxing, Pong from the NaturalEnv collection  \citep{zhang2018natural}, which allow background customization with solid colors; the actions are game-specific.
In the Breakout environment, scores typically range between 0 and 200-300, representing a satisfactory final score. In Boxing, scores fall within the range of [-100, 100], where a score of 100 indicates that the agent defeats the opponent without sustaining any hits. For Pong, scores range from [-21, 21], with 21 signifying victory over the opponent without conceding a single point.
As we did for the CarRacing environment, we first evaluate end-to-end performance of standard and R3L models followed by the stitching evaluation.

\subsubsection{End-to-end performance}\label{appendix:end-to-end}
Results can be seen in \Cref{tab:atari-no_stitching}. Performance of R3L models for Pong and Boxing are comparable to those of the absolute models. Breakout, however, has much lower scores. We attribute this to the higher visual complexity caused by the numerous bricks in the level.

\begin{table}[ht]
\caption{Episode mean scores for models trained end-to-end, therefore when no stitching is performed. Models trained with (\textit{R3L}) have comparable performance, with performance loss only on the Breakout environment.}
\label{tab:atari-no_stitching}
\centering
\footnotesize
\begin{tabular}{ccccc}
\toprule
 & & & \texttt{Visual variations} & \\
& & \texttt{plain} & \texttt{green} & \texttt{red} \\
\midrule
\multirow{2}{*}{\rotatebox{0}{\textbf{Pong}}} & \textit{E. Abs} & 21 $\pm$ 0 & 21 $\pm$ 0 & 21 $\pm$ 0 \\
& \textit{E. R3L} (ours) & 21 $\pm$ 0 & 20 $\pm$ 1 & 21 $\pm$ 0 \\
\cmidrule{2-5}
\multirow{2}{*}{\rotatebox{0}{\textbf{Boxing}}} & \textit{E. Abs} & 95 $\pm$ 2 & 95 $\pm$ 3 & 96 $\pm$ 2 \\
& \textit{E. R3L} (ours) & 95 $\pm$ 3 & 93 $\pm$ 4 & 88 $\pm$ 6 \\
\cmidrule{2-5}
\multirow{2}{*}{\rotatebox{0}{\textbf{Breakout}}} & \textit{E. Abs} & 298 $\pm$ 63 & 262 $\pm$ 61 & 132 $\pm$ 19 \\
& \textit{E. R3L} (ours) & 146 $\pm$ 60 & 77 $\pm$ 25 & 119 $\pm$ 135 \\
\bottomrule
\end{tabular}
\end{table}

\subsection{Zero-shot stitching}\label{appendix:stitching-atari}
\paragraph{Zero-shot stitching performance}
\Cref{tab:atari-stitching_performance} presents zero-shot stitching evaluation. Although \textit{S. R3L} models outperform absolute ones in all the environments, there is a significant performance drop in the mean performance indicated by the lower mean scores, with respect to \textit{E. R3L}. The high the standard deviation, however, signifies that some of the models are still able to perform well in some cases.
We argue that the high precision required by Atari games might be the reason for the performance drop in stitching, as even small noise in the encoders' latent spaces can bring to minor mistakes in action predictions, which in turn can bring to a losing condition in the game. Meanwhile, the CarRacing environment is far more accommodating, and in the event of a mistake, the policy can compensate in subsequent actions.

\begin{table}[h!]
    \caption{Comparing R3L and absolute in zero-shot stitching performance.
    The original domains for the encoders and the controllers are listed in the rows and columns, respectively.
    Although performing worse than what done in the Carracing environment, R3L still outperforms the absolute model in all the tests.}
    \label{tab:atari-stitching_performance}
    \resizebox{\textwidth}{!}{
    \begin{tabular}{cccccccccccccc}
    \toprule
    & & & \multicolumn{6}{c}{\textbf{Controller}} & & \\
    \cmidrule{4-12}
    & & & \multicolumn{3}{c}{\textbf{Pong}} & \multicolumn{3}{c}{\textbf{Boxing}} & \multicolumn{3}{c}{\textbf{Breakout}} \\
    \cmidrule(r){4-6} \cmidrule(l){7-9} \cmidrule(l){10-12}
    & & & \texttt{plain} & \texttt{green} & \texttt{red} & \texttt{plain} & \texttt{green} & \texttt{red} & \texttt{plain} & \texttt{green} & \texttt{red} \\
    \cmidrule(r){1-6} \cmidrule(l){7-9} \cmidrule(l){10-12}
    \multirow{9}{*}{\rotatebox{90}{\textbf{Encoder}}} 
    & \multirow{2}{*}{\rotatebox{90}{\texttt{plain}}} 
    & \emph{S. Abs} & -21 $\pm$ 0 & -21 $\pm$ 0 & -21 $\pm$ 1 & -29 $\pm$ 10 & -25 $\pm$ 22 & -33 $\pm$ 5 &  8 $\pm$ 6 & 12 $\pm$ 5 & 8 $\pm$ 5 \\
    & & \emph{S. R3L} & $\mathbf{0} \pm 20$ & $\mathbf{-2} \pm 19$ & $\mathbf{7} \pm 17$ & $\mathbf{65} \pm 39$ & $\mathbf{11} \pm 54$ & $\mathbf{46} \pm 40$ & $\mathbf{71} \pm 73$ & $\mathbf{16} \pm 18$ & $\mathbf{17} \pm 11$ \\[1.5ex]
    & \multirow{2}{*}{\rotatebox{90}{\texttt{green}}} 
    & \emph{S. Abs} & -20 $\pm$ 1 & -21 $\pm$ 0 & -21 $\pm$ 0  & -18 $\pm$ 17 & -21 $\pm$ 15 & -33 $\pm$ 10 &  9 $\pm$ 7 & 15 $\pm$ 4 & 7 $\pm$ 6 \\
    & & \emph{S. R3L} & $\mathbf{-1} \pm 20$ & $\mathbf{-7} \pm 19$ & $\mathbf{8} \pm 18$ & $\mathbf{66} \pm 38$ & $\mathbf{42} \pm 36$ & $\mathbf{58} \pm 27$ & $\mathbf{28} \pm 19$ & $\mathbf{64} \pm 100$ & $\mathbf{44} \pm 59$ \\[1.5ex]
    & \multirow{2}{*}{\rotatebox{90}{\texttt{red}}} 
    & \emph{S. Abs} & -21 $\pm$ 0 & -21 $\pm$ 0 & -21 $\pm$ 0 & -27 $\pm$ 17 & -20 $\pm$ 16 & -38 $\pm$ 13 &  9 $\pm$ 6 & 13 $\pm$ 5 & 8 $\pm$ 5 \\
    & & \emph{S. R3L} & $\mathbf{1} \pm 18$ & $\mathbf{-3} \pm 18$ & $\mathbf{6} \pm 20$  & $\mathbf{62} \pm 40$ & $\mathbf{20} \pm 56$ & $\mathbf{49} \pm 39$ & $\mathbf{17} \pm 9$ & $\mathbf{25} \pm 28$ & $\mathbf{14} \pm 10$ \\[1.5ex]
    \bottomrule
    \end{tabular}
        }
\end{table}

\subsection{Computing}
We trained all of our models on an RTX 3080 with an Intel i7-9700K CPU @ 3.60GHz and 64 GB of RAM.
All of our models were trained with the CleanRL implementation of PPO, running 16 environments in parallel.
\paragraph{CarRacing}
We trained all of our CarRacing models for a total of 5 million steps. Training took around 3h for every model, except for the slow task, which took 4, mainly because episodes were set to be longer to allow the car to complete the track, and therefore evaluation episodes took longer to complete.

\paragraph{Atari Suite}
Games in the Atari suite were trained for 10 million steps. Pong and Breakout required around 2h30m, boxing 2h40m.
\end{document}